\documentclass[manuscript,screen]{acmart}
\usepackage{csquotes}
\usepackage{multirow}
\usepackage{caption}
\usepackage{subcaption}
\usepackage{xurl}

\AtBeginDocument{%
  \providecommand\BibTeX{{%
    \normalfont B\kern-0.5em{\scshape i\kern-0.25em b}\kern-0.8em\TeX}}}

\setcopyright{acmcopyright}
\copyrightyear{2022}
\acmYear{2022}

\acmConference[ACM FAccT 2022]{ACM FAccT 2022: ACM Conference on Fairness, Accountability, and Transparency}{June 21--24, 2022}{Seoul, South Korea}
\acmBooktitle{ACM FAccT 2022: ACM Conference on Fairness, Accountability, and Transparency,
  June 21--24, 2022, Seoul, South Korea}

\usepackage{savesym}

\usepackage{comment}
\savesymbol{comment}
\excludecomment{mycomment}

\usepackage{changes}

\begin{document}

\title{Seeing without Looking: Analysis Pipeline for Child Sexual Abuse Datasets}


\author{Camila Laranjeira}
\affiliation{
  \institution{Department of Computer Science, Universidade Federal de Minas Gerais}
  \streetaddress{Av. Pres. Antônio Carlos, 6627}
  \city{Belo Horizonte}
  \state{Minas Gerais}
  \country{Brazil}
  \postcode{31270-901}
 }

\author{João Macedo}
\affiliation{
  \institution{Department of Computer Science, Universidade Federal de Minas Gerais}
  \streetaddress{}
  \city{Belo Horizonte}
  \state{Minas Gerais}
  \country{Brazil}
  \postcode{}
}

\author{Sandra Avila}
\affiliation{%
  \institution{Artificial Intelligence Lab.
(Recod.ai), Institute of Computing, University of Campinas}
  \streetaddress{Cidade Universitária Zeferino Vaz}
  \city{Campinas}
  \country{Brazil}
  \postcode{13083-970}
 }

\author{Jefersson A. dos Santos}
\affiliation{
  \institution{Department of Computer Science, Universidade Federal de Minas Gerais}
  \streetaddress{Av. Pres. Antônio Carlos, 6627}
  \city{Belo Horizonte}
  \state{Minas Gerais}
  \country{Brazil}
  \postcode{31270-901}
}
\email{camilalaranjeira@ufmg.br, joão, sandra@ic.unicamp.br,jefersson@dcc.ufmg.br}








\begin{abstract}
    The online sharing and viewing of Child Sexual Abuse Material (CSAM) are growing fast, such that human experts can no longer handle the manual inspection. However, the automatic classification of CSAM is a challenging field of research, largely due to the inaccessibility of target data that is — and should forever be — private and in sole possession of law enforcement agencies. To aid researchers in drawing insights from unseen data and safely providing further understanding of CSAM images, we propose an analysis template that goes beyond the statistics of the dataset and respective labels. It focuses on the extraction of automatic signals, provided both by pre-trained machine learning models, e.g., object categories and pornography detection, as well as image metrics such as luminance and sharpness. Only aggregated statistics of sparse signals are provided to guarantee the anonymity of children and adolescents victimized. The pipeline allows filtering the data by applying thresholds to each specified signal and provides the distribution of such signals within the subset, correlations between signals, as well as a bias evaluation. We demonstrated our proposal on the Region-based annotated Child Pornography Dataset (RCPD)
    , one of the few CSAM benchmarks in the literature, composed of over 2000 samples among regular and CSAM images, produced in partnership with Brazil's Federal Police. Although noisy and limited in several senses, we argue that automatic signals can highlight important aspects of the overall distribution of data, which is valuable for databases that can not be disclosed. Our goal is to safely publicize the characteristics of CSAM datasets, encouraging researchers to join the field and perhaps other institutions to provide similar reports on their benchmarks.
\end{abstract}


\maketitle

\section{Introduction}

Child Sexual Abuse (CSA) is one of the major issues we are currently tackling as a society. Its mitigation is listed as one of the 17 Global Goals for Sustainable Development outlined in 2015 at the United Nations General Assembly\footnote{\url{https://www.globalgoals.org}}. The abuse can take many forms, it may involve physical contact or violent acts~\cite{mmfdh2021}, but it may also consist in the online sharing of child images for sexual purposes, to which we refer as Child Sexual Abuse Material (CSAM). The latter is growing exponentially, according to reports from the National Center for Missing and Exploited Children (NCMEC)~\cite{bursztein2019rethinking}. The same report shows that the generation of novel content is also rapidly increasing, and its nature varies widely, from violent acts such as organized groups abducting children to abuse, document, and later share with online communities~\cite{silbergcase} to the massive number of families innocently sharing pictures of their children on social media, which sex offenders later download to compose their gallery~\cite{maxim2016online}.  

CSAM can be defined as a type of media portraying children that may or may not be involved in sexually inciting situations, but used by adults for sexual purposes. The nomenclature adopted throughout the literature may vary from Child Pornography (CP)~\cite{yiallourou2017detection} to Child Exploitation Material (CEM)~\cite{dalins2018laying} or even Indecent Images of Children (IIOC)~\cite{kloess2019challenges}. The Luxembourg Guidelines\footnote{\url{https://ecpat.org/luxembourg-guidelines}} currently used by many law enforcement agencies, including Interpol\footnote{\url{https://www.interpol.int/Crimes/Crimes-against-children/Appropriate-terminology}}, provides clear indications of appropriate terminologies, using the terms Sexual Abuse and Sexual Exploitation to address the severity of the content. Those terms also aid in defining a clear distinction between criminal acts against children and pornography/indecency, with the latter being associated with adult content that may be consensual and overall legal. Since sexual exploitation has specific connotations of profit and exchanges~\cite{unicef2020a}, we adopt the broader term Child Sexual Abuse Material. In our work, we focus on the domain of images. 


Methodologies for CSAM detection commonly resort to two types of data: real evidence from law enforcement investigations or legal images acquired through photo libraries or search engines. Some approaches only use legal images, attempting to solve subtasks from the target problem, for instance, performing age estimation and pornography detection~\cite{macedo2018benchmark, gangwar2021attm}. Nonetheless, CSA datasets are still required to validate the solution in real scenarios. A similar version of the following disclaimer can be found in any paper handling CSAM, and it applies to our work as well:

\begin{displayquote}
Due to the sensitive nature of our research, we are working in partnership with the Brazilian federal law enforcement agency, whose experts are the only ones with direct access to any sensitive data mentioned throughout this paper.
\end{displayquote}

That disclaimer leads to the main motivation for our research. Since CSA images are illegal for civilians to possess and share, every single dataset in the literature is --- and should forever be --- private and in sole possession of law enforcement agencies. Whenever such data is used for training machine learning solutions, models are usually private as well, which is essential to protect children and adolescents involved in the research and avoid future exploitation. 


The need for privacy, although essential, leads to disadvantages for researchers, law enforcement, and ultimately society as a whole. CSAM detection is a scarcely researched field partly due to data inaccessibility; hence, studying it is challenging throughout the entire research process: from proposing a methodology without ever seeing the data to evaluating and drawing conclusions from results. Therefore, scientific contributions in this field are somewhat limited since algorithms and models can not be easily scrutinized or subject to further inspection. Additionally, it is difficult to compare results with other works from literature since researchers usually refer to local partnerships with law enforcement agencies, and there is no established benchmark worldwide.

Furthermore, classification labels for CSAM are inherently ambiguous, both when defining what is sexually explicit and whether the depicted person is, in fact, a child. According to \citet{kloess2019challenges}, the question is challenging even for law enforcement experts. Authors found high disagreement among experts when labeling pictures regarding indecency and age groups, the latter especially difficult for older adolescents. The work in \cite{hessick2016refining} cites the six-factor ``Dost test'', a set of rules to qualify an image or video as child abuse. Not only it was considered a vague and broad definition, but it may also cause further harm to the victims as it encourages looking for subjective sexual cues from the children. In practice, law enforcement experts often refer to the context in which the picture was found, such as data from the investigation or other pictures from the same device. Although the domain does not allow unambiguous classification, researchers can contribute by investigating a wide range of visual cues that may be relevant in the production of triage tools and priority queues to ease the burden of law enforcement agents.

Another aspect worth highlighting is the lack of information regarding types of bias that might exist in CSAM databases. There are known tendencies in reported sexual abuse cases, especially in demographic dimensions such as gender, age, and race, with the most common victims in Brazil being black and brown girls from 8 to 14 years old~\cite{mmfdh2019}. But, as far as our knowledge goes, there are no reports on bias and tendencies of apprehended CSAM on a content level, with most reports limited to traffic data and volume of apprehensions~\cite{bursztein2019rethinking}. 

Works in the literature attempt to circumvent both the issue of label ambiguity and lack of content-level information. The works in \cite{macedo2018benchmark, dalins2018laying} provide more thorough labels for images, such as bounding boxes for nude body parts or objective classification labels of whether people are actively engaging in sexual activities. However, they rely on the laborious process of assigning manual labels. Our goal is to investigate how to safely publicize the characteristics of CSA datasets without adding to the burden of law enforcement experts. In this work, we assess the validity of extracting automatic signals from a real CSA database to provide a comprehensive documentation. We do not advocate for publicizing information on individual samples. Instead, we evaluate how aggregated statistics on the entire dataset or even disaggregated by specific subsets can be the source of valuable insights.

Our focus is on defining a set of attributes relevant to the target domain. For that purpose, we inspect the literature on CSAM detection and reports from law enforcement institutions to select the features of interest for which automatic labeling is viable. Then, inspired by Dataset Nutrition Labels~\cite{holland2020dataset} and tools like Google Know Your Data~\cite{kyd2021}, we define a set of visualizations and metrics from which we can draw insights on each attribute as well as relations between them. Since the target domain does not allow for widely sharing individualized attributes extracted from the database, we focus on defining a set of visualizations that allows a comprehensive inspection of the source data. To assess the validity of our proposal, we apply it to the 
Region-based annotated Child Pornography Dataset (RCPD)~\cite{macedo2018benchmark}, a benchmark produced by the Federal Police of Brazil, due to its extensive set of labels. This work is a step towards a future product, awaiting ethics reviews and law enforcement assent, of a freely available interactive tool for researchers to explore RCPD
's characteristics beyond the aspects presented in this paper, and hopefully new CSA benchmarks in the future.

Results can be used to support empirical claims from the literature, such as the tendency of CSAM happening in indoor environments~\cite{yiallourou2017detection}. We also found that CSAM apprehended in Brazil has different tendencies than reports of sexual abuse with physical contact, since RCPD 
is overwhelmingly composed of light-skinned individuals. Moreover, the research itself surfaced gaps in the literature that might contribute to the field of CSAM detection, for instance, the benefits of an object detection approach focusing on child-related objects and reinforcing the need for better age estimation models. We hope that results will instigate other researchers to join the field and encourage dataset owners to provide similar documentation to their benchmarks.

\section{Related Work}

This section is divided into two main topics of interest. First, we explore the literature on CSAM detection to draw insights from data and visual features commonly used as resources for training and validation. Then, we focus on the importance of inspecting datasets presenting theoretical research and practical templates previously proposed.

\subsection{CSAM Detection}

In the early years of research and applications on child sexual abuse image detection, researchers, law enforcement, and other institutions relied mostly on hash-based approaches. With comprehensive databases such as the one provided by the National Center for Missing \& Exploited Children (NCMEC)\footnote{\url{https://www.missingkids.org}} 
and hash-based methodologies such as PhotoDNA~\cite{dna2008}, one can perform an effective search for previously reported instances of CSAM. However, with the exponential growth of novel content generation~\cite{bursztein2019rethinking}, content-based approaches to classify previously unseen images are increasingly essential. Although there are important contributions in the literature leveraging filenames~\cite{panchenko2012detection}, network attributes~\cite{shupo2006toward}, and even folder structure~\cite{guerra2021detecting}, our work focuses on visual cues that indicate the presence of CSA.

For the last decade, the literature on CSAM detection evolved little regarding the type of semantic information modeled, with most efforts dedicated to improving the applied techniques. Early approaches such as NuDetective~\cite{de2010nudetective} and iCOP~\cite{peersman2014icop} rely on image descriptors crafted to capture nudity, while \citet{sae2014towards}'s work also added a child classification stream based on texture and distances between facial landmarks to distinguish adult from child nudity. 

Current approaches usually leverage deep learning techniques, which are more robust and achieve better accuracy scores than earlier works, but the goal remains roughly the same: detecting children and nudity/pornography. Some approaches tackle both tasks. The works of \cite{macedo2018benchmark, rondeau2019deep}, for instance, rely on Yahoo OpenNSFW model~\cite{mahadeokar2016open} to detect pornography cues, and propose their age estimation approaches. While \citet{rondeau2019deep} leverages label distributions to assess apparent age, \citet{macedo2018benchmark} propose a single-model estimation of child presence, age, and gender. A more recent work~\cite{gangwar2021attm} explores a wide range of technical improvements over neural networks such as residual connections along with inception and attention modules to propose separate models for age estimation and pornography detection. \citet{gangwar2021attm} also propose Juvenile-80k, gathering around 24 thousand underage images from a wide range of age estimation datasets and supplementing it with images crawled from public search engines. It is important to note that this type of collection does not abide by ethical standards such as UNICEF's Responsible Data for Children report~\cite{young2019responsible}, but it indicates an important gap in the literature: age estimation models specialized in children and adolescents. 

There is a growing body of research focusing solely on age estimation in the context of CSAM, and the collection of child images seems to be a common choice for many. \citet{anda2020deepuage} propose a model specialized in underage individuals, building a novel dataset for age and gender estimation, VisAGe. They gathered over 19 thousand faces of individuals under 18 years old, from creative-commons licensed Flickr images. Similar to \citet{gangwar2021attm}, \citet{castrillon2018evaluation} and \citet{chaves2020improving}  gather images from several age estimation databases, amounting to a large number of underage samples, with no extra supplementation of data. \citet{castrillon2018evaluation} propose AgeMega, a dataset with over 15 thousand underage faces and around 30 thousand adult ones. They explore a multitude of local descriptors to train support vector machines along with convolutional neural networks (CNN) predictions, to compose a score-level fusion approach as the final classification method. \citet{chaves2020improving} apply synthetic eye occlusions to a portion of samples, based on the assumption that criminals can do the same to omit the identities of victims and fine-tune a neural network for age estimation in that scenario.

Much of the work in CSAM detection is focused on engineering features or proposing models based on similar prior definitions of relevant attributes. One work that differs in that sense is \cite{vitorino2018leveraging}, in which authors directly model the target task, proposing an end-to-end binary classification approach fine-tuned on real CSA data. However, a broader investigation focusing on understanding different aspects of the data is essential. The works of \cite{yiallourou2017detection, kloess2019challenges} are among the few providing valuable insights on a wider range of attributes. \citet{yiallourou2017detection} propose a synthetic dataset associating levels of appropriateness to images in terms of a variety of features, such as gestures, scene type, illumination, and facial expressions. The work of \cite{kloess2019challenges}, on the other hand, does not attempt automatic classification, but draws insights from human experts, highlighting visual cues that may cause or solve ambiguities in classification.

In the literature, there is little concern with providing statistical measures on CSA data and extracted attributes, regarding their occurrence and correlation. In this work, we not only bring a wide perspective on relevant features for CSAM detection as \cite{yiallourou2017detection,kloess2019challenges}, but we also provide an analysis framework to compose a comprehensive documentation of datasets and demonstrate it in a real benchmark available for testing.   

\subsection{Dataset Inspection and Documentation} \label{sec:data-analysis}


An extensive inspection of source data should be the first step of a machine learning solution. \citet{sambasivan2021everyone} highlight the negligence of both researchers and companies in analyzing and documenting datasets. Their main focus are databases used for training, concluding that poor quality data has significant downstream effects on trained models and ultimately may compromise the validity of results. However, inspection and documentation is just as crucial for test benchmarks. \citet{narayanan2021ethics} explores how test benchmarks are a central guide to model selection for practitioners. In this sense, poor data may lead to poor solutions being widely adopted as state-of-the-art. 

Machine learning researchers are only recently adhering to the practice of documentation as a response to the growing number of works proposing specific guidelines. Some propositions suggest verbal descriptions regarding aspects like the data's origin, structure, collection process, recommended applications, among many others. Datasheets for Datasets~\cite{gebru2018datasheets} is among the most complete propositions in that sense, defining an extensive set of questions researchers and practitioners should reply to when proposing or releasing a dataset. In specific domains such as Natural Language Processing, we also find templates such as the data cards currently used for datasets on Hugging Face~\cite{mcmillan2021reusable}.  

Other works tend to approach documentation as a mix of verbal descriptions and attributes statistics, the latter through visualizations or summaries. For instance, \citet{prabhu2020large} inspect large-scale image datasets, presenting a dataset audit card for ImageNet to display how sensitive attributes automatically extracted or manually labeled are distributed. Nutrition Labels~\cite{holland2020dataset}, on the other hand, is a more robust proposition inspired by food labels to describe ``ingredients'' of a dataset. Adding summary statistics and pair plots for all variables in the dataset allows a comprehensive inspection of the data, not only as a means to understand it but also to draw insights from it. 

There is a myriad of visualization tools allowing dataset inspection. We highlight a specific feature from Google Know Your Data~\cite{kyd2021}. Besides general statistics presented as histograms, the ``relations'' tab implements a fairness metric proposed in \cite{aka2021measuring} to assess fairness without labels as a measure of normalized correlation. In our work, the presentation of attributes follows propositions from both Know Your Data and Nutrition Labels. We suggest investing more heavily in relations between attributes. Since we are working with automatically extracted signals representing a wide variety of semantic information, the goal is to assess how they can be relevant for tasks related to CSAM.





\section{Methodology}

Our goal is to safely publicize characteristics of child sexual abuse image datasets aimed at researchers willing to contribute to the field. There are a couple of priorities to keep in mind. We should avoid revictimizing children and adolescents depicted in the images. Preserving their anonymity is the main priority, as well as preventing any attempts to content reproduction. Thus, we do not expose dense features extracted from the samples and advise researchers in the field to do the same. Since deep learning approaches are becoming so efficient at generating synthetic content, dense features from CSA images might be misused for such purposes. Our pipeline is solely based on sparse annotations previously provided by dataset owners and sparse automatic signals extracted from images, mainly composed of classification and detection labels along with metrics to estimate characteristics like image quality or skin tone.

The first challenge of our work is choosing which attributes might be relevant for a CSA database. We resort to the literature from both computer science researchers attempting to tackle the problem of CSAM detection and reports from law enforcement agents and entities regarding relevant aspects when searching for CSAM in an apprehension. 

Since we do not intend to release data on individual samples, an additional challenge arises: how to present the aggregated attributes in a useful manner, allowing researchers to explore the characteristics of the database. Section~\ref{sec:analysis} describes the proposed approach. This work is a step towards a future product, awaiting ethics reviews and law enforcement assent, of an interactive tool that will allow researchers to explore RCPD
's characteristics way beyond the aspects presented in this paper. Thus, some visualization resources we mention may rely on user~interactivity.



\subsection{Attributes}\label{sec:att}

The vast majority of recent approaches to CSAM detection divide the problem into two subtasks: age estimation and pornography classification, since those are the closest domains to the target task with large availability of data. However, as concluded by \citet{kloess2019challenges}, forensic experts usually rely on much more than that to classify images, especially for ambiguous samples. For instance, subjects interviewed by the authors reported that the environment provides valuable cues, both in terms of scene-related features (e.g., outdoor vs. indoor) and object information (e.g., indications of child-like environments due to the presence of toys). \citet{yiallourou2017detection} go a step further, modeling not only the aforementioned features but also aspects like illumination, classifying darker scenes as more suspicious. 

An additional aspect considered for choosing which attributes were relevant to our proposal is to unveil potential sources of bias in the data. According to a 2019 report from the Equipe da Ouvidoria Nacional dos Direitos Humanos, a Brazilian public agency for human rights, sexual abuse and exploitation of children are strongly biased towards gender, age, and race, with the most reported victims being black and brown girls in their late childhood/early adolescence (8 to 14 years old)~\cite{mmfdh2019}. The importance of those demographic attributes is also highlighted in the work of \cite{macedo2018benchmark} by the authors choice of including those labels in their proposed benchmark, RCPD
. Later, in Section~\ref{sec:subdem}, we discuss the advantages and disadvantages of automatically extracting demographic attributes. 

Considering both aspects, discriminative features and potential sources of bias, all features leveraged by our proposal are summarized in Table \ref{tab:att}, along with the respective attributes derived from them. Table \ref{tab:att} also divides attributes into per individual versus per sample, meaning a single sample can have multiple instances of the same attribute associated with it. Although each attribute will be specified in the remaining of this section, with a discussion on its relevance, we can anticipate a few types of recurring attributes. For instance, we collect the output \textit{probabilities} of any given model and derive the \textit{class} inferred according to a specified threshold. That allows users to input their desired threshold after looking at probability distributions and update the class counts' aggregated information. Visualizing probabilities with the ranking of classes also provides insights into potential noises and uncertainties. 

We can also highlight the \textit{number of instances} associated with features that may occur more than once in an image and represent either an overall count of occurrences or a per class count. Finally, the \textit{standard deviation} is calculated for demographic attributes, providing insights on aspects such as age difference or skin tone diversity per sample.


\begin{table}[t]
\small
\begin{tabular}{p{0.13\linewidth}p{0.22\linewidth}p{0.26\linewidth}p{0.29\linewidth}}
\\ \toprule
&  & \textbf{Per individual} & \textbf{Per sample} \\ \hline
Labels&$*$&$*$&$*$ \\ \hline
\multirow{5}{*}{Demographics}&Face~\cite{zhang2016joint}                & probability, absolute area, relative area & class, \#instances \\
&ITA~\cite{kinyanjui2020fairness}                 & average value & standard deviation   \\
&Child~\cite{macedo2018benchmark}               & probability, class & \#instances ($n_c$), standard deviation \\
&Age~\cite{macedo2018benchmark}                 & probabilities ($n_c$), class & \#instances ($n_c$), standard deviation  \\ 
&Gender presentation~\cite{macedo2018benchmark} & probability, class & \#instances ($n_c$), standard deviation \\ \hline
\multirow{2}{*}{Pornography}&NSFW~\cite{mahadeokar2016open} & & probability, class   \\
&Pornography~\cite{man}         &  & probabilities ($n_c$), class \\ \hline
\multirow{2}{*}{Context}&Objects~\cite{bochkovskiy2020yolov4}             & class ($2$), absolute area, relative area  & \#instances ($n_c + 1$)  \\ 
&Scenes~\cite{zhou2017places}              &   & class ($3$)  \\ \hline
\multirow{2}{*}{Quality}&Luminance           &  & average value  \\
&BRISQUE~\cite{piq}             & & value  \\ \hline
\multirow{4}{*}{Metadata}&File extension      &   & value   \\
&Colormode           &   & value   \\
&Aspect ratio        &   & value   \\
&Image resolution    &   & value   \\ \bottomrule 
\end{tabular}
\vspace{1mm}
\caption{Attributes derived from each extracted feature. Numbers and variables in parenthesis are added to instances that can be derived into multiple attributes, with $n_c$ representing the number of classes of the respective feature. Since labels depend on the evaluated dataset, we added a placeholder variable with asterisks where attributes would be listed.}
\label{tab:att}
\end{table}


Although this work largely focuses on the advantages of automatic signals extracted from samples, our proposal also aims at providing a deeper understanding of labels produced by dataset owners. Thus, the experiments section will provide a complete description of attributes related to labels. Table~\ref{tab:att} contains a placeholder variable entitled ``Labels'' to indicate that the derived attributes and proposed visualizations will also apply to original information from the dataset.

\subsubsection{Demographics}\label{sec:subdem}

Extracting automatic signals on demographics is a highly sensitive decision, since automatic models are extremely limited in their ability to model such complex features as gender, race, and even age. Available models refer to gender as a binary classification task, while we can hardly call race a classifiable concept, especially for countries like Brazil with a wide range of phenotypes in its population. Regarding age, although it can be classified from visual features to a certain extent, in the context of CSAM, there is a crucial confusion boundary in the range of adolescence to early adulthood. Findings in \cite{rosenbloom2013inaccuracy} show that even medical experts in pediatric and adolescent development show a high error rate despite the availability of maturity cues such as face, breasts, body contour, and pubic area. The subjects classified two out of three images of young-looking adult women as adolescents.

On the other hand, reports of child sexual abuse involving physical contact are highly biased towards specific demographics~\cite{mmfdh2019}, but little is known of how such attributes are distributed in materials shared online. Disaggregating CSA data, as well as CSAM classification, by demographic dimensions is thus essential. Unfortunately, that is not a domain in which we can acquire self-reported demographics, since in most cases, the identities of victimized children and perpetrators depicted in the images are unknown to law enforcement. Thus, even if demographic information is labeled, it can only be a subjective view from labelers. Even so, such annotations are costly to produce, and most CSA databases do not provide them. For large-scale databases that may arise in the future, which unfortunately is viable due to the number of images and videos apprehended by law enforcement, annotating demographics may be impracticable. 

\begin{figure}[t]
    \centering
    \includegraphics[width=0.65\textwidth]{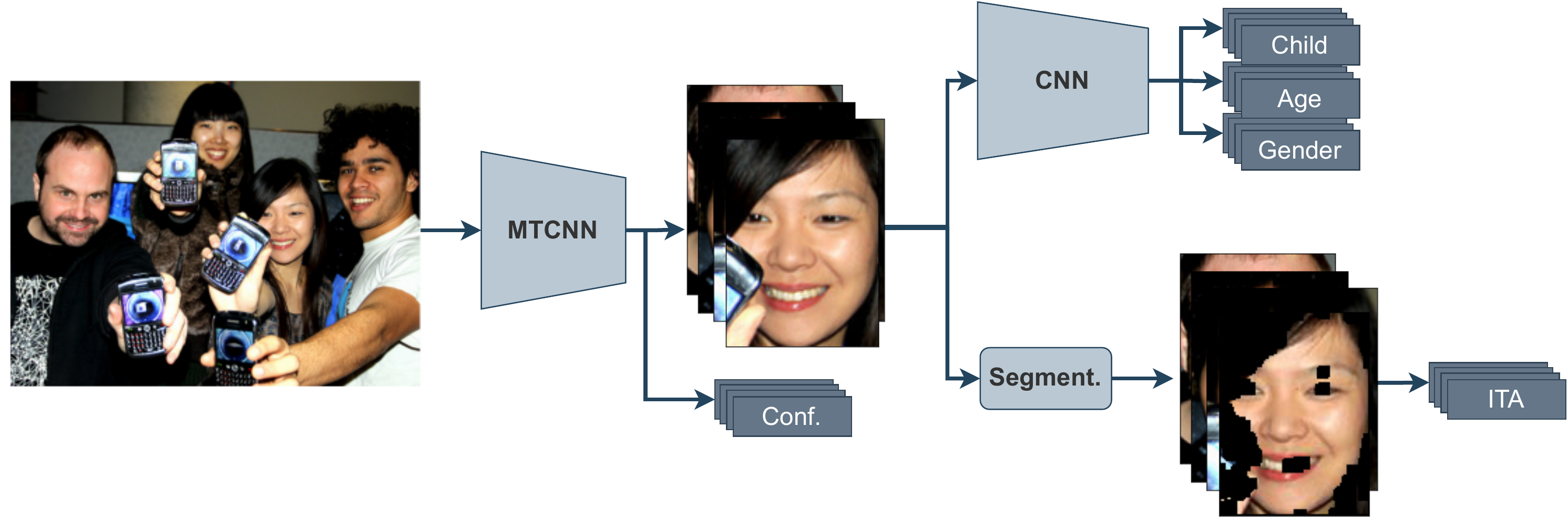}
    \caption{Pipeline to extract demographic features. Trapezoids represent neural networks, rounded rectangles are functions, and darker rectangles are outputs. MTCCN refers to multi-task cascaded CNN~\cite{zhang2016joint} and ITA refers to individual typology angle. Input photo retrieved from Open Images database~\cite{kuznetsova2018open}.}
    \label{fig:demographics}
\end{figure}

We argue it is important to leverage automatic features on demographics to view the general tendency of a CSA image database. In the context of our work, the bottom line is not to classify individual samples but rather to get a general view of a group of samples. Fig.~\ref{fig:demographics} summarizes the collection process of demographic features. Apart from skin tone, all features we leverage are commonly estimated from facial images, evoking the need for a face detection module, which by itself generates relevant attributes (refer to Table \ref{tab:att}). Many references in the literature rely on MTCNN~\cite{zhang2016joint} as the chosen tool for face detection~\cite{macedo2018benchmark, chaves2020improving, gangwar2021attm} since it is one of the most accurate and robust in the literature. Therefore, it was the model of our choice as well. The extracted face then serves as input for subsequent procedures.

\citet{macedo2018benchmark} propose a single model to perform three tasks. First, a binary classification of whether the input face belongs to a child. Secondly, it estimates the age-group of a subject among the categories of Adience~\cite{eidinger2014age}: 0-2, 4-6, 8-13, 15-20, 25-32, 38-43, 48-53, 60+. Note that there is little concern with discerning underage individuals (assuming Brazil's legal age of 18), but as we have argued before, CSAM detection should be limited to its role as a triage tool, and ambiguities can be solved by human experts. Finally, the model provides an estimate of binary gender presentation. Although there are methods with greater accuracy for age estimation, the joint prediction of age and child performed in \cite{macedo2018benchmark}, achieving an accuracy of $94\%$ on the second task, produces an age classifier less skewed towards adulthood.   

Lastly, there is skin tone estimation. \citet{kinyanjui2020fairness}, a work in the field of skin lesion analysis, relied on the metric entitled Individual Typology Angle (ITA), which was found to be strongly correlated with the Melanin Index~\cite{wilkes2015fitzpatrick} and can be quantitatively measured from each pixel in an image, according to the following equation:
\begin{equation*}
    ITA = \frac{arctan(L - 50)}{b} \times \frac{180}{\pi},
\end{equation*}

\noindent where $L$ and $b$ are channels of an image in CIE-Lab space, respectively indicating luminance and amount of yellow. As in \cite{kinyanjui2020fairness}, the final score is an average of ITA measures within one standard deviation of the distribution. However, different from skin lesion datasets, our data is not comprised solely of naked skin. The work of \cite{merler2019diversity} proposes a selection of skin regions based on facial landmarks. However, we found that approach works best with front-facing samples. So, we decided to apply a simple skin segmentation algorithm, still limiting it to facial images to minimize background clutter or even the presence of clothes, which would require a more robust approach. The segmentation algorithm was adapted from a public project on Github~\cite{Jean2018}, based on the watershed region-growing approach~\cite{vincent1989morphological}. Input markers are defined by adding the output from explicit boundaries of skin regions in two color spaces, HSV (H $<25$, S $>40$)~\cite{tsekeridou1998facial} and YCbCr ($77<$ Cb $<127$, $133<$ Cr $<173$)~\cite{chai1999face}, followed by morphological operations of erosion and dilation to discard~noise.      

\subsubsection{Pornography}

A comparative evaluation on pornography detection methods~\cite{gangwar2017pornography} found that Yahoo's OpenNSFW model~\cite{mahadeokar2016open} achieved the best accuracy, over $87\%$, in a CSA database provided by the Spanish Police, despite it being a model trained solely on adult pornography. \citet{gangwar2021attm} found that combining OpenNSFW with an age estimation model achieves over $83\%$ accuracy for CSAM detection in more challenging settings, including adult pornography in the test database. Although their proposed model outperforms OpenNSFW, neither the model nor the dataset used for training are yet released to the research community. \citet{macedo2018benchmark} also relied on OpenNSFW to incorporate their detection pipeline, achieving superior performance over existing CSAM detection approaches. 

In this work, we use OpenNSFW as one of the two approaches for pornography detection. But since it only provides binary labels, we also experiment with an open-source project that tackles a 5-category classification task, estimating probabilities for drawing, hentai, neutral, porn, sexy~\cite{man}. Although there is no cartoon-related data on the datasets used in this work, we were interested in a more fine-grained distinction provided by the categories porn and sexy. 

\subsubsection{Context}

We could not find any approach for CSAM detection tested against real-world data, which uses the context information in the form of scene and object features to perform the target task. The closest reference is \cite{yiallourou2017detection}, in which authors hand-labeled a synthetic dataset with binary labels for indoor/outdoor environment and presence/absence of what they considered suspicious objects. As previously mentioned, insights from forensic experts revealed that context can be valuable for disambiguation of samples~\cite{kloess2019challenges}. Thus, we experiment with well-established approaches for both scene classification and object detection tasks.

Regarding objects, YOLO, currently in its 4th version~\cite{bochkovskiy2020yolov4}, is by far one of the best in the literature both in terms of accuracy and time performance. All attributes regarding objects are derived from YOLOv4 pre-trained on COCO~\cite{lin2014microsoft}, meaning it is able to estimate probabilities for 80 object classes, which are hierarchically organized into 12 macro classes: person, furniture, indoor, kitchen, electronic, animal, vehicle, food, appliance, sports, accessory, and outdoor. Thus, for each sample, we derive two class attributes from YOLO: base-level class and macro-level class. Regarding the \textit{number of instances}, we derive a count for the overall presence of objects and the count of instances for each base-level object category. We are especially interested in the person category, since detecting people beyond just faces is valuable in the context of CSAM, as faces can be absent or occluded.

A recent survey on scene classification~\cite{zeng2021deep} found that VGG-Places~\cite{zhou2017places}, a VGG architecture trained on Places365, is still competitive with single model approaches relying on global features from the input. Patch-based approaches or even ensemble methods, which tend to be more computationally demanding, can achieve higher accuracy on known benchmarks. We opted to favor VGG-Places, a lighter model, as a proof of concept. In the Discussion and Future Directions section, we bring the topic of scene classification models tailored for the domain of CSAM.    

\subsubsection{Quality}

Following insights provided in \cite{yiallourou2017detection}, we were interested in the relevance of quality assessment metrics to define what they called appropriateness of an image. Thus, we extracted both the average luminance of images in CIE-Lab space and BRISQUE, a no-reference approach to estimate image quality, provided by a Pytorch framework~\cite{piq}.


\subsubsection{Metadata}

The choice of including basic file information has two main reasons. First, those are fundamental for low-level implementation choices, such as reshaping input images. Secondly, they are cheap to generate, leading to a favorable cost-benefit relationship. We followed the procedure presented in Google's Know Your Data tool~\cite{kyd2021}, extracting the following information: file extension, image color mode, aspect ratio, and resolution.


\subsection{Presentation}\label{sec:analysis}

Planning how the attributes are presented is essential to maximize the level of inspection allowed, especially since the source data can not be seen. As Dataset Nutrition Labels~\cite{holland2020dataset} and tools such as Know Your Data~\cite{kyd2021}, we provide summary statistics and relations among attributes, divided into the following types:


\begin{itemize}
    \item \textbf{General Distributions}: Histograms represent rankings, discrete attributes, and multimodal distributions of continuous attributes (e.g., probabilities). For continuous unimodal distributions, boxplots are used instead. 
    \item \textbf{Disaggregated Distributions}: The same visualizations provided by general distributions can be disaggregated by up to 3 attributes, leveraging facet plots and color-coding distributions.    
    \item \textbf{Co-occurrence}: Heatmaps with simultaneous occurrences of pairs of attributes as raw or normalized counts.     
    \item \textbf{Correlation}: {Heatmap visualization of correlation. As described in \cite{aka2021measuring}, Point-wise Mutual Information normalized by $P(y)$ estimates the correlation between pairs of variables w.r.t chance, defined as
    \begin{equation*}
        nPMI_y = \Big( ln\frac{P(x,y)}{P(x) \cdot P(y)} \Big) \Big/ -ln~P(y), 
    \end{equation*}
    with $X$ and $Y$ being the two variables (attributes) compared. An additional information is included in this visualization, as implemented by \cite{kyd2021}, providing the ratio between real $C_{x,y}$ and expected $\Tilde{C}_{x,y}$ co-occurrence matrices. Expectation is defined as the mutual information from independent marginal probabilities. Consider $c$ the sum over all values of $C_{x,y}$, expected values are as follows: 
    \begin{equation*}
        \Tilde{C}_{x,y} = c * (P(x) \cdot P(y)).
    \end{equation*}
    The color-coding of heatmap cells represents the ratio between real and expected co-occurrence, and it is only visually depicted if the discrepancy exceeds a $95\%$ confidence interval. 
    }
\end{itemize}

To produce co-occurrence and correlation matrices and to disaggregate distributions for all attributes, numeric variables need to be quantized. For values referring to probabilities, quantization is performed by uniformly dividing data into intervals of $0.1$, producing 10 bins in total. The remaining variables are also divided into 10 uniform bins, lower bound to $max(min_x,~\overline{x} - 1.96\sigma)$ and upper bound to $min(max_x,~\overline{x} + 1.96\sigma)$, with $\overline{x}$ and $\sigma$ representing the distribution's average and standard deviation. The bins at both extremes comprise all values below/above them.

\subsection{Research Ethics}


The study in its entirety had the participation of an expert from the Federal Police of Brazil, the sole responsible for handling any sensitive data mentioned throughout this paper. To ensure data integrity, random spot checks were conducted by this officer during the extraction of automatic signals, to confirm the validity of attributes and correspondence to the referred sample. We assured original media and individual data points would never leave the police's servers, thus sharing with the authors of this paper only visualizations of aggregated statistics. 



Data collected for each instance is anonymized, providing no sensitive information on victimized children, perpetrators, or law enforcement entities. Additionally, we do not intend to publicize individualized attributes, only visualizations of aggregated data, making it even more difficult to expose individual samples.  
We are still waiting for an official assessment regarding the release of a public visualization tool to allow user interactivity, which will only be made effective after approvals from both law enforcement and the ethics board from Universidade Federal de Minas Gerais.



\section{Case Study: RCPD
}

The goal of our experiments is to validate the relevance of automatic signals, specifically the aforementioned attributes, to extract valuable knowledge from real CSAM data. Additionally, we assess whether the previously outlined visualization resources can provide comprehensive understanding without releasing individual data points. To do so, we derive all attributes from a benchmark entitled region-based annotated child pornography dataset (RCPD)
~\cite{macedo2018benchmark}, with 2138 samples among CSAM, adult pornography, and non-sensitive images, although there are no hard labels for such categories. It was proposed as a robust benchmark for testing CSAM classification approaches and associated tasks, with annotations of body parts bounding boxes (person, head, breast/chest, genital, and buttocks), along with subjective labels for demographic attributes (age, gender, and ethnicity). From those base-level labels, other attributes are derived, such as nudity level corresponding to the exposition of body parts (none, seminude, nude, sex), and binary CSAM labels are provided as a combination of nudity level and perceived age, providing researchers with the freedom to classify CSAM by different thresholds if relevant. The original work considers children to be under 13 years old, and sensitive images to contain at least one person fully nude or engaging in sexual activity. By that criteria, RCPD provides 836 CSAM samples and 285 depictions of adult pornography. Appendix \ref{app:nutrition} provides a complete description of attributes and their summary statistics heavily inspired by the work in \cite{holland2020dataset}, which proposes a Nutrition Label for datasets.  

The choice of RCPD
to validate our proposal was driven by the extensive annotations provided by the authors, thus assuring that 
patterns and relationships found by automatic signals are not artificially produced by prediction errors. An exhaustive report of all possible settings would be impracticable since most visualizations are relations between pairs of attributes. Therefore, the remaining of this section leverages different combinations of attributes and visualizations to highlight relevant aspects of the extracted attributes. Appendix \ref{app:vissummary} depicts a more thorough set of visualizations for labels and attributes, highlighting the potential of releasing a tool to the research community for independent studies.


\begin{figure}
     \centering
     \begin{subfigure}[b]{0.9\textwidth}
         \centering
         \includegraphics[width=\textwidth]{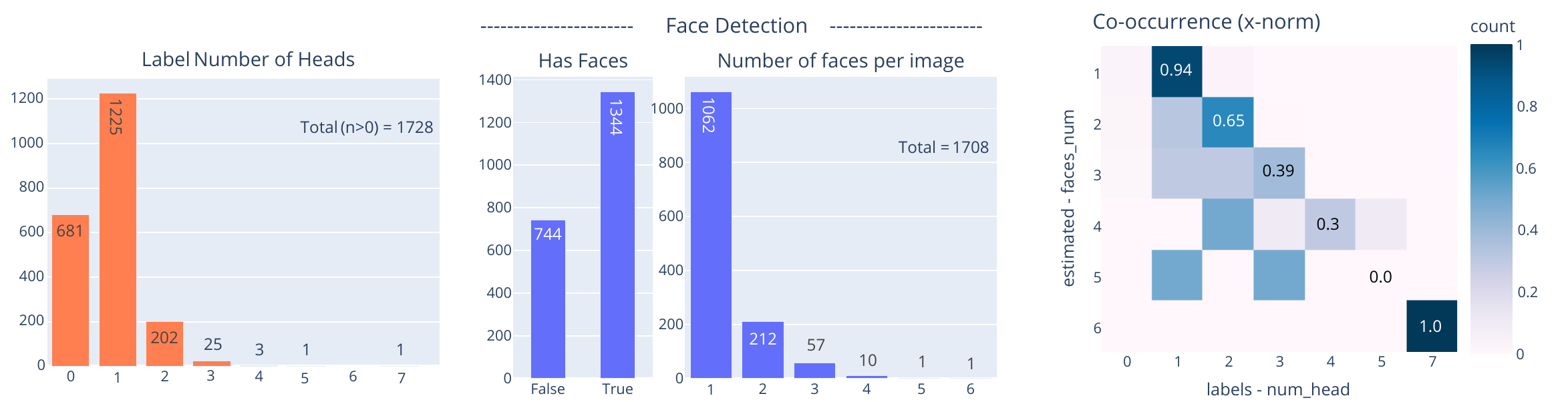}
         \caption{Face detection}
         \label{fig:face-detection}
     \end{subfigure}
     \\
     \begin{subfigure}[b]{0.9\textwidth}
         \centering
         \includegraphics[width=\textwidth]{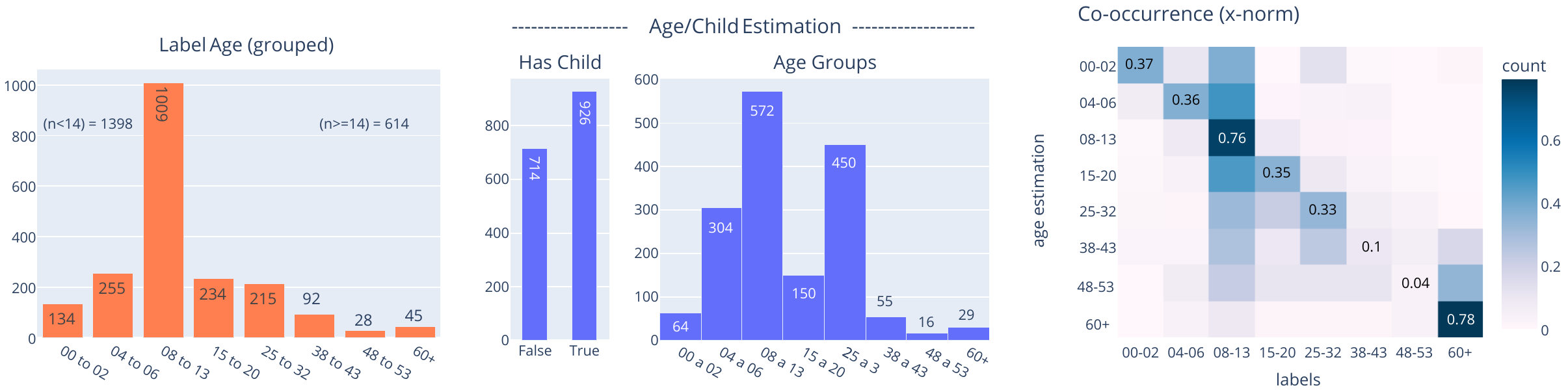}
         \caption{Age estimation}
         \label{fig:age-estimation}
     \end{subfigure}
     \\
     \begin{subfigure}[b]{0.45\textwidth}
         \centering
         \includegraphics[width=\textwidth]{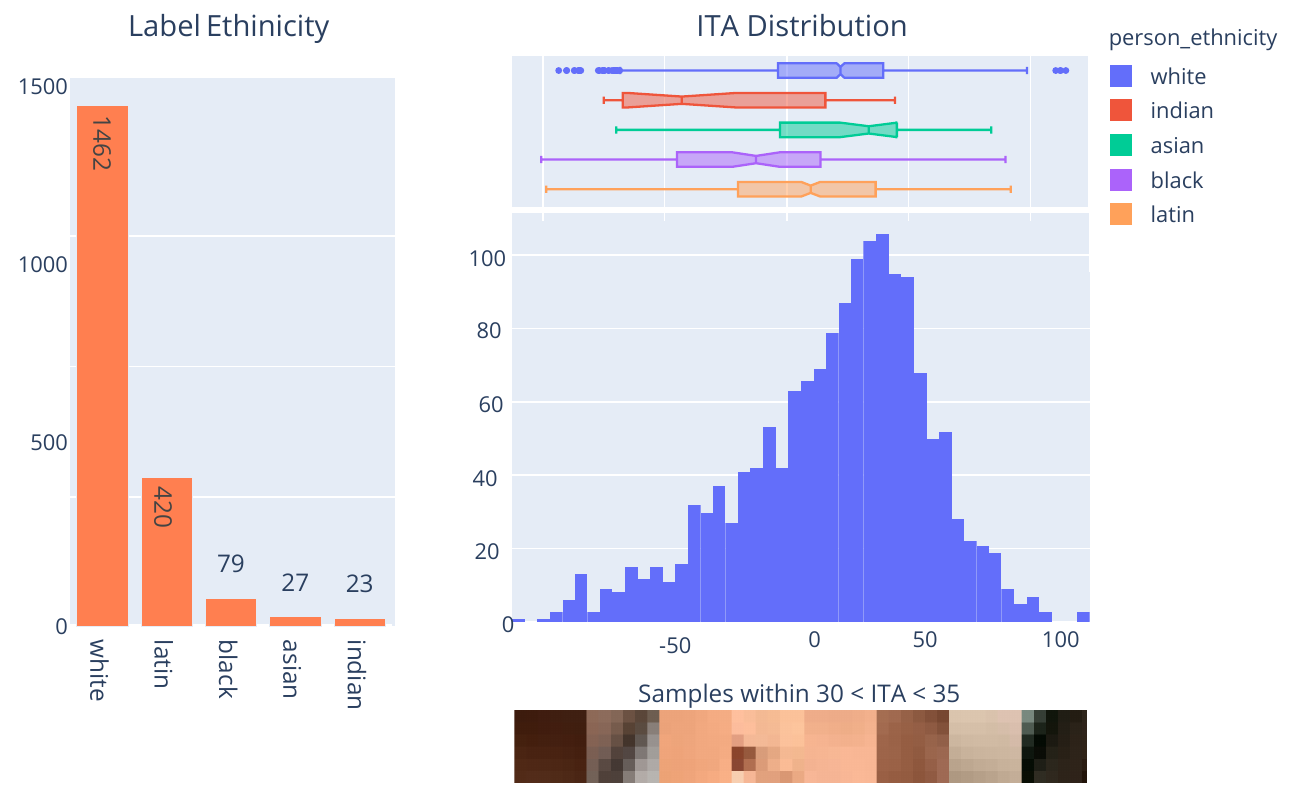}
         \caption{Perceived ethnicity and ITA}
         \label{fig:ethnicity}
     \end{subfigure}
     \begin{subfigure}[b]{0.4\textwidth}
         \centering
         \includegraphics[width=\textwidth]{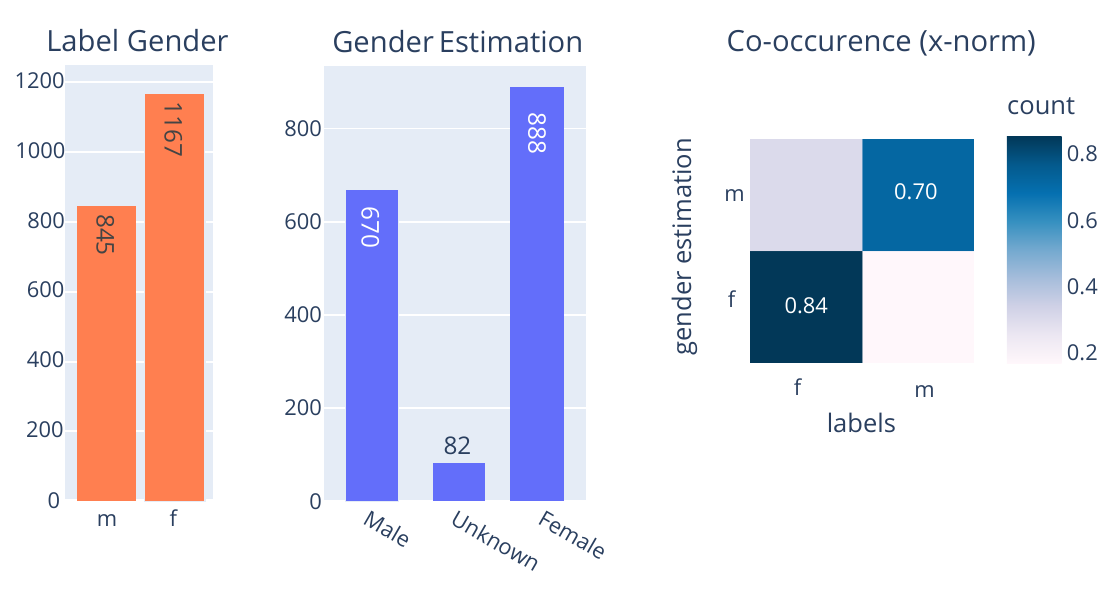}
         \caption{Perceived gender}
         \label{fig:gender}
     \end{subfigure}
        \caption{Labels and features extracted for demographic attributes. Label distribution in orange and on the left of each subfigure, features extracted in purple and in the middle, and co-occurrence of signals to the right, except for ethnicity and ITA, in which marginal distributions are disaggregated by labels for ethnicity.}
        \label{fig:graphics-demographics}
\end{figure}

First, let us look at attributes that directly relate to labels in the dataset. Fig.~\ref{fig:graphics-demographics} compares demographic attributes automatically extracted with the respective labels. Notably, the database is highly skewed regarding age (Fig.~\ref{fig:age-estimation}) and ethnicity (Fig.~\ref{fig:ethnicity}), depicting mostly white children under 15 years old. Although black and brown children are the main target of sexual abuse in Brazil, according to forensic experts most CSAM apprehended in Brazil appears foreign in nature, which might explain the overwhelming number of samples depicting white children. Even though ITA can not be used to classify ethnicity effectively~\cite{karkkainen2019fairface}, its distribution makes it clear the predominance of light-skinned individuals. As we did not wish to classify skin tones as it is usually done in research for skin lesion classification~\cite{kinyanjui2020fairness}, we opted to add interactivity so that users can click on any bin to see $6\times6$ skin patches from the database associated with its values. Fig.~\ref{fig:ethnicity} demonstrates it by showing a few instances for the average ITA.    

Regarding age, we stress the importance of better age estimation methods focusing on underage individuals. Although the method we chose had this concern in mind, and it was able to detect the predominance of children around 8-13 years old, the results still overestimate the presence of young adults. We highlight that age estimation is still dependent on face detection, which by itself is not perfect. But RCPD 
has over 170 samples of children not showing face~\cite{macedo2018benchmark}, which means they would go unnoticed by current automatic methods. Finally, gender classification captures the skewness towards the female gender, consistent with labels provided by RCPD 
and overall reports on child abuse. 

\begin{figure}
     \centering
     \begin{subfigure}[b]{0.15\textwidth}
         \centering
         \includegraphics[width=\textwidth]{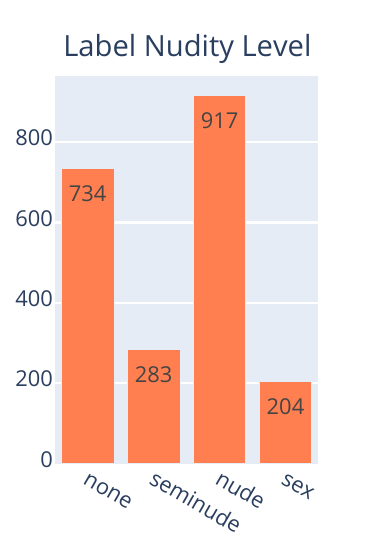}
         \caption{Label: Nudity}
         \label{fig:y equals x}
     \end{subfigure}
     \begin{subfigure}[b]{0.3\textwidth}
         \centering
         \includegraphics[width=\textwidth]{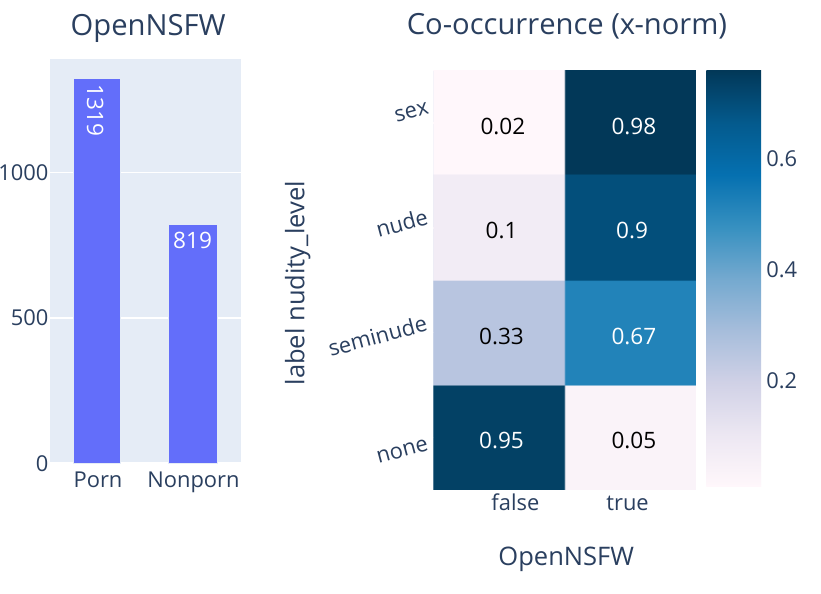}
         \caption{Yahoo OpenNSFW predictions}
         \label{fig:nsfw}
     \end{subfigure}
     \begin{subfigure}[b]{0.4\textwidth}
         \centering
         \includegraphics[width=\textwidth]{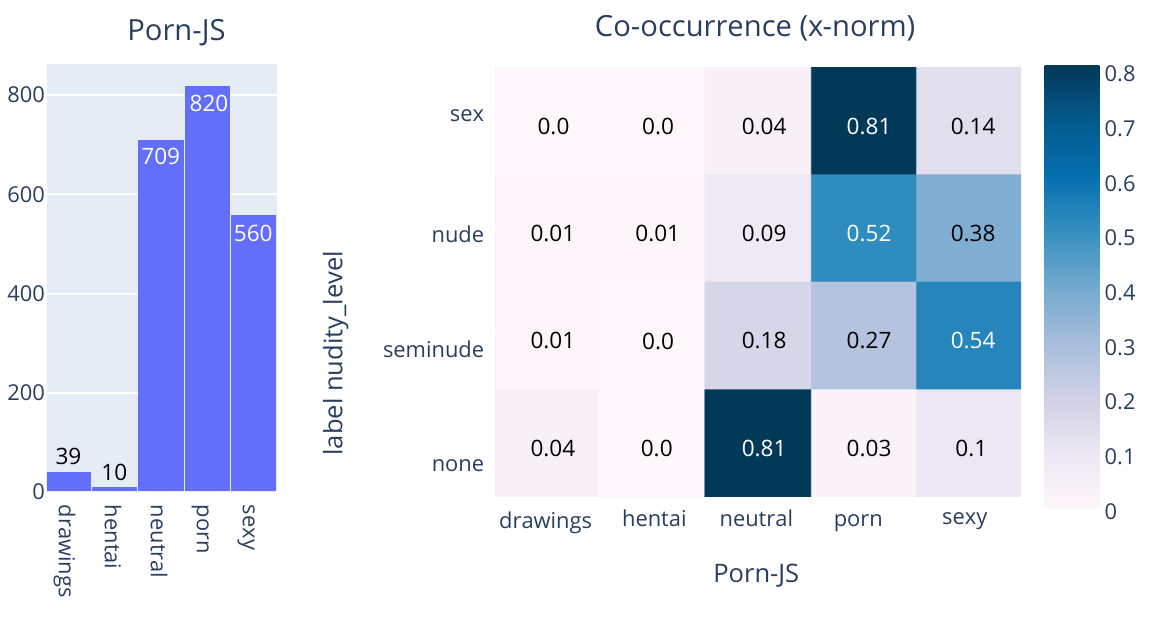}
         \caption{Porn-JS predictions}
         \label{fig:js}
     \end{subfigure}
        \caption{Labels and features extracted for pornography attributes. We provide the predictions for both classification methods: Yahoo OpenNSFW and Porn-JS, as well as co-occurrences with original labels.}
        \label{fig:pornography}
\end{figure}

Pornography classification can be directly related to labels for nudity level, as shown in Fig.~\ref{fig:pornography}. We set a fairly low probability threshold $t=0.3$ for Yahoo's OpenNSFW~\cite{mahadeokar2016open}, the same protocol adopted in~\cite{macedo2018benchmark}. Although it is over $90\%$ accurate in detecting nude and sex samples (Fig.~\ref{fig:nsfw}), they are both associated with the porn category. Porn-JS~\cite{man} achieves roughly the same accuracy on nude and sex samples, with a more fine-grained approach, assigning two levels of sensitivity (porn and sex), while being more sensitive to seminude instances (Fig.~\ref{fig:js}). Further studies are required to evaluate if the same can be achieved by defining threshold intervals to the binary classification of OpenNSFW.

\begin{figure}
    \centering
    \begin{subfigure}[b]{0.49\textwidth}
        \centering
        \includegraphics[width=\textwidth]{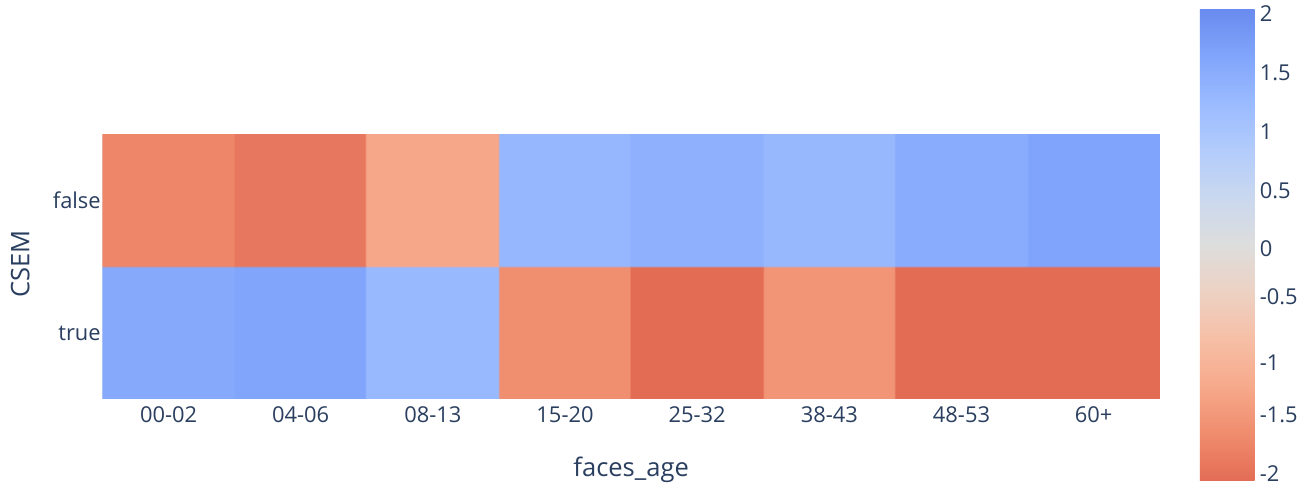}
         \caption{Age vs. CSAM}
         \label{fig:y equals x}
    \end{subfigure}
    \begin{subfigure}[b]{0.49\textwidth}
        \centering
        \includegraphics[width=\textwidth]{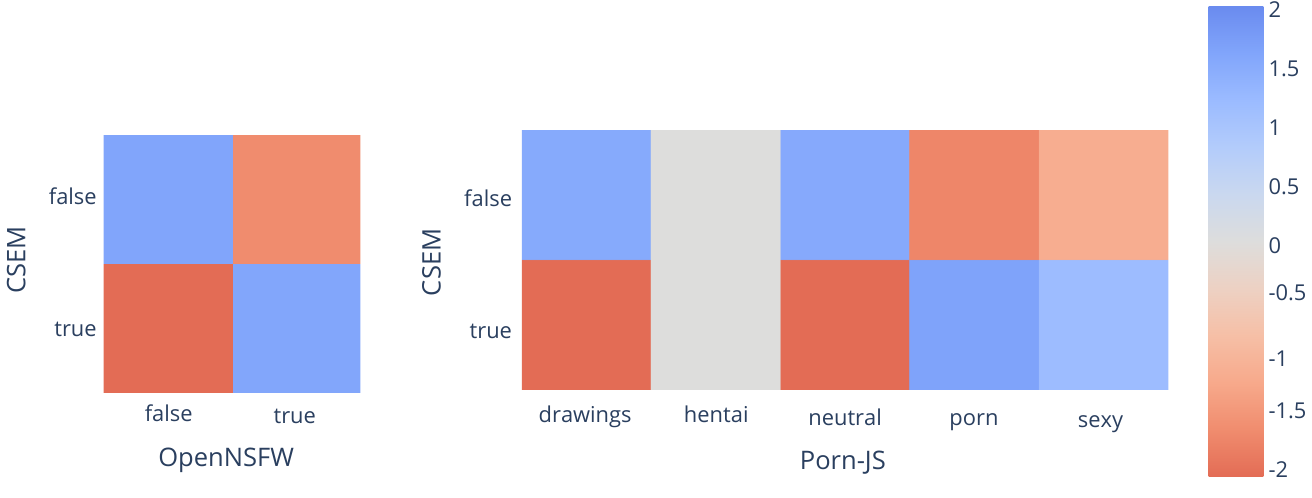}
         \caption{Pornography vs. CSAM}
         \label{fig:y equals x}
    \end{subfigure}
    \caption{Correlation between pornography classification methods, age estimation and CSAM categories.}
    \label{fig:porn-age-pmi}
\end{figure}

One aspect worth highlighting regarding age and pornography is the prevalence of categories relative to CSAM labels. Fig.~\ref{fig:porn-age-pmi} shows correlation metrics, indicating an over-representation of pornographic samples in the CSAM category, and a prevalence of children under 14. Logically, those are the two attributes contemplated by the category's definition. However, to make RCPD 
more challenging, it is worth balancing samples for those attributes as best as possible, especially since the distinction among adult pornography, CSAM, and safe children photos is so critical to the field.

\begin{figure}
    \centering
    \includegraphics[width=0.95\textwidth]{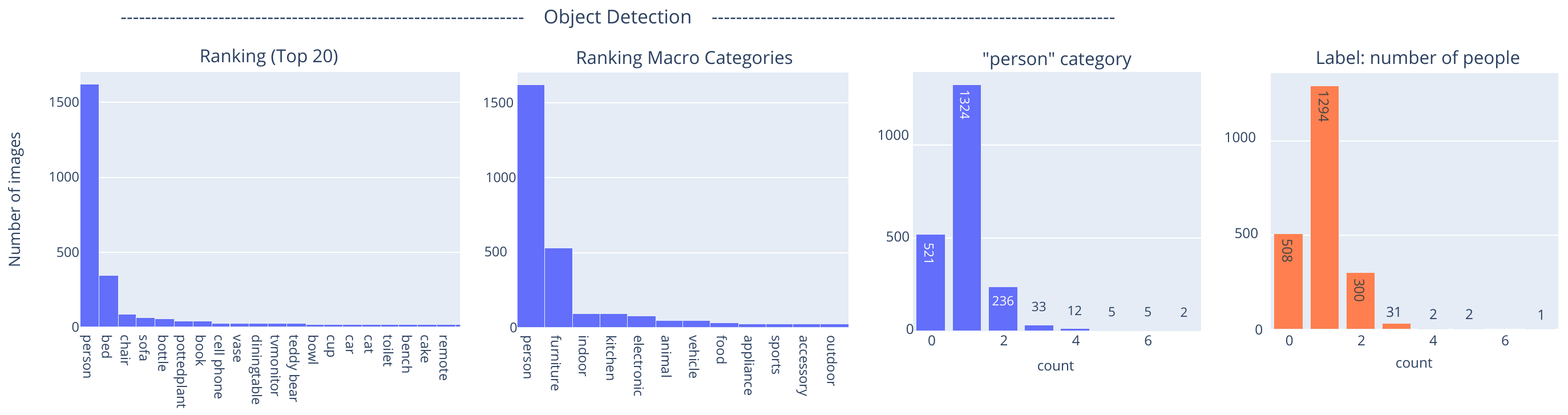}
    \caption{Rankings of object detection base-level and macro-level categories, along with distributions for the ``person'' category and correspondent labels from region-based annotated child pornography dataset (RCPD)
    .}
    \label{fig:object-person}
\end{figure}

As for context-related features, Fig.~\ref{fig:object-person} shows that RCPD 
is a highly person-centric dataset, with furniture as the second most common macro-category of objects, indicating a prevalence of indoor environments as scene classification will later confirm. The ``person'' category can be directly related to labels from RCPD 
regarding number of people per sample. The two graphs on the right of Fig.~\ref{fig:object-person} show that object detection is able to capture an accurate distribution of people per sample, showing that the majority of images on RCPD 
is comprised of a single person.    


\begin{figure}
    \centering
    \includegraphics[width=0.85\textwidth]{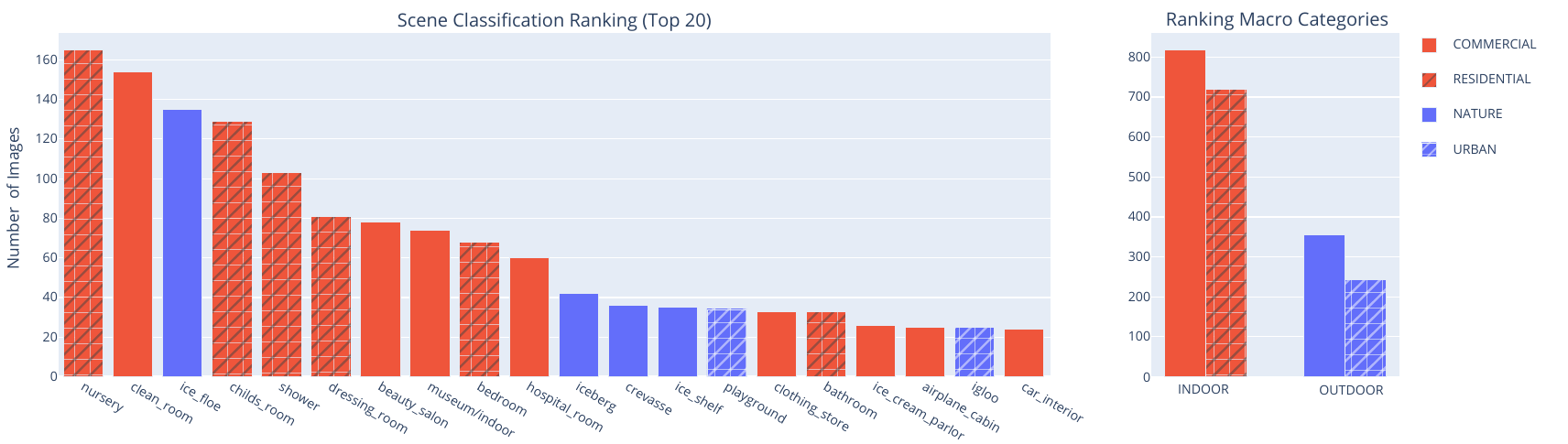}
    \caption{Scene classification Ranking for base-level and macro categories.}
    \label{fig:scenes}
\end{figure}

Scene classification is hierarchically organized in three levels, as shown in Fig.~\ref{fig:scenes}. For RCPD
, most samples are indoor, as object categories previously indicated. A qualitative assessment showed that some base-level classification instances that appear to be predominant, such as ``clean\_room'' and ``ice\_floe'', are actually misclassifications. On the other hand, we see the predominance of child-related categories, such as ``nursery'', ``childs\_room'', and ``playground''. It is worth specializing scene classification approaches for the domain of CSAM, with fewer and domain-driven categories, since law enforcement experts usually search for context cues such as a child-like environment or public places~\cite{kloess2019challenges}. 

\begin{figure}
    \centering
    \includegraphics[width=0.95\textwidth]{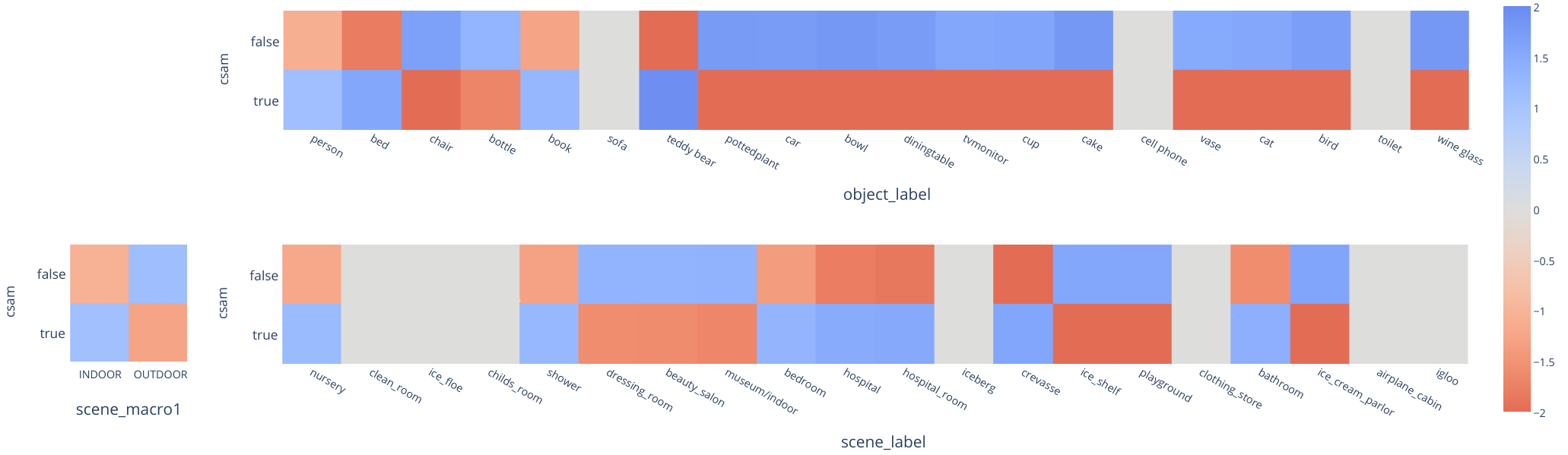}
    \caption{Correlation between scenes, objects\added{,} and CSAM categories.}
    \label{fig:scenes-cor}
\end{figure}

Looking at how context features correlate with CSAM labels in Fig.~\ref{fig:scenes-cor}, most residential categories are positively correlated with CSAM. Since data collection was not concerned with balancing context, it indicates that looking for residential cues in images might benefit law enforcers as an additional triage dimension. From a macro perspective, results add more evidence to support the statement from \cite{yiallourou2017detection} that CSAM occurs more often in indoor environments. Regarding object information, COCO classes do not explore child-like categories more extensively, but we could see that class ``teddy bear'' is somewhat predominant and strongly correlated with CSAM. This tendency could be weakened if RCPD 
is ever balanced for the presence of children, but it brings valuable insight into the importance of contextual cues.

\section{Discussion and Future Directions}

Our work has two main focuses. First and foremost, we explore the literature searching for a set of characteristics that can be automatically extracted and relevant to CSAM-related tasks. Afterward, we investigate how to give safe publicity to such attributes. The automatic extraction of features has the downside of relying on measures that are uncertain and prone to errors; however, it produces cheap annotations without adding to the burden of law enforcement agents providing benchmarks for the research community. By evaluating our proposal on RCPD
, we count on a vast set of labels to show that automatic features are reliable for analyzing general tendencies in a group of samples while also surfacing novel and valuable insights. At the same time, presenting attributes as aggregated statistics complies with the goal of safe publicity since it does not expose individual samples. \vspace{1mm}

\noindent \textbf{Expanding the boundaries of CSAM detection.} The literature on CSAM detection usually focuses on attributes related to age and pornography, but forensic experts provide valuable insights on a wide variety of visual cues that can be further explored in future research. For instance, we showed that context information from scene and object classification are correlated with CSAM labels, highlighting that residential scenes and child-like visual cues often occur in scenes of child abuse; hence they are valuable dimensions for detection. Notably, no reference in the literature focusing on contextual cues in the domain of CSAM. Methods such as object detection specialized in child-like instances, or scene classification with fewer and domain-centric categories could be valuable to the field. 

We can think of further information relevant to CSAM detection. For instance, including sexual organ detection, as recently proposed by \citet{tabone2021pornographic}, which was already applied in the context of CSAM detection in \cite{al2020evaluating}. As the labels of RCPD 
show, such information is relevant for automatic triage of forensic samples. A couple of works also mention facial expressions as a relevant cue for disambiguation in CSAM detection~\cite{yiallourou2017detection, kloess2019challenges}, since children presenting apparent discomfort or unhappiness could be under a stressful situation, perhaps being forced to pose for a picture. \vspace{1mm}

\noindent \textbf{On biases in CSAM datasets.}
We chose some of our attributes driven by a crucial lack in the literature since little is known about biases in CSAM. First, regarding demographic dimensions, which are essential to produce fair machine learning solutions, RCPD 
indicates that CSA data shared online has a different distribution from reported cases of physical abuse in Brazil in terms of race, with RCPD  
overwhelmingly composed of light-skinned individuals. On the other hand, both domains indicate that most victimized children are girls within a range of 8 to 13 years old. Thus, aggregated accuracy measures for CSAM detection may hide performance discrepancies regarding sensitive attributes. We argue that in terms of cost-benefit, automatic labels have the advantage of easily allowing disaggregated inspections, as they can surface large performance discrepancies among subgroups.        
 
Concerning biases, we can assess how challenging and adequate a dataset can be for training and evaluating models. One of the main challenges for CSAM is distinguishing it from legal images of children and adult pornography; thus, it is essential to balance available benchmarks in both dimensions. For RCPD
, there is statistically significant correlation between age-groups and CSAM categories, as well as pornography levels and CSAM, suggesting room for improvement. \vspace{1mm}



\noindent \textbf{Safe publicity to CSAM documentations.}
It is easy to understand why researchers provide little to no descriptions on the content of child sexual abuse material, being it a highly sensitive domain. However, a proper evaluation of machine learning methods in terms of accuracy and fairness requires knowing the data to a certain extent. CSAM detection is very challenging in terms of reproducibility and comparison of results; thus, researchers should invest in providing the characteristics of the data used for training or validation if it is unknown to the community. We explore a range of documentation practices in the literature, showing that extracting sparse attributes and presenting them as aggregated statistics is both valuable and safe. Since we do not intend to release individual data points, all reports remain anonymous. Additionally, deriving each feature into multiple attributes and investing more heavily in relations between attributes allows to potentially produce thousands of visualizations surfacing different aspects of the data. \vspace{1mm}



\noindent \textbf{Future directions.}
This work is a step towards a future product to allow independent inspections from researchers willing to join the field. We wish to produce a freely available interactive tool with attributes from RCPD 
and all visualization capabilities explored throughout this paper. Such a tool can be expanded to accept predictions from researchers who submit their evaluation methods on the respective benchmark. It would allow authors to scrutinize their proposition beyond aggregated measures of accuracy provided in leaderboards, to assess opportunities for improvement of their approaches. Since this is a high-stakes domain, in which the downstream application is finding evidence of child sexual offenses, law enforcement oversight on the development and critical use of such tools are essential. This work is the first step in a long endeavor towards a more transparent and still safe field of CSAM detection. 

\section*{Acknowledgements}

This work is partially supported by Serrapilheira Institute under grant Serra–R-2011-37776. The authors also acknowledge the support from FAPEMIG under Grant APQ-00449-17, along with CNPq under grants 311395/2018-0 and 424700/2018-2, and CAPES under Finance Code 001. Sandra Avila is partially funded by CNPq PQ-2 (315231/2020-3), FAPESP (2013/08293-7, 2020/09838-0), H.IAAC (Artificial Intelligence and Cognitive Architectures Hub), and Google LARA 2021. None of the funding sources had any role in the design and conduct of this study.

\bibliographystyle{ACM-Reference-Format}
\bibliography{main}

\appendix

\newpage
\section{RCPD 
Nutrition Labels}\label{app:nutrition}
This section provides brief descriptions of attributes contained in RCPD 
along with summary statistics following Nutrition Labels guidelines~\cite{holland2020dataset}.

\begin{figure}[H]
    \centering
    \includegraphics[width=0.88\textwidth]{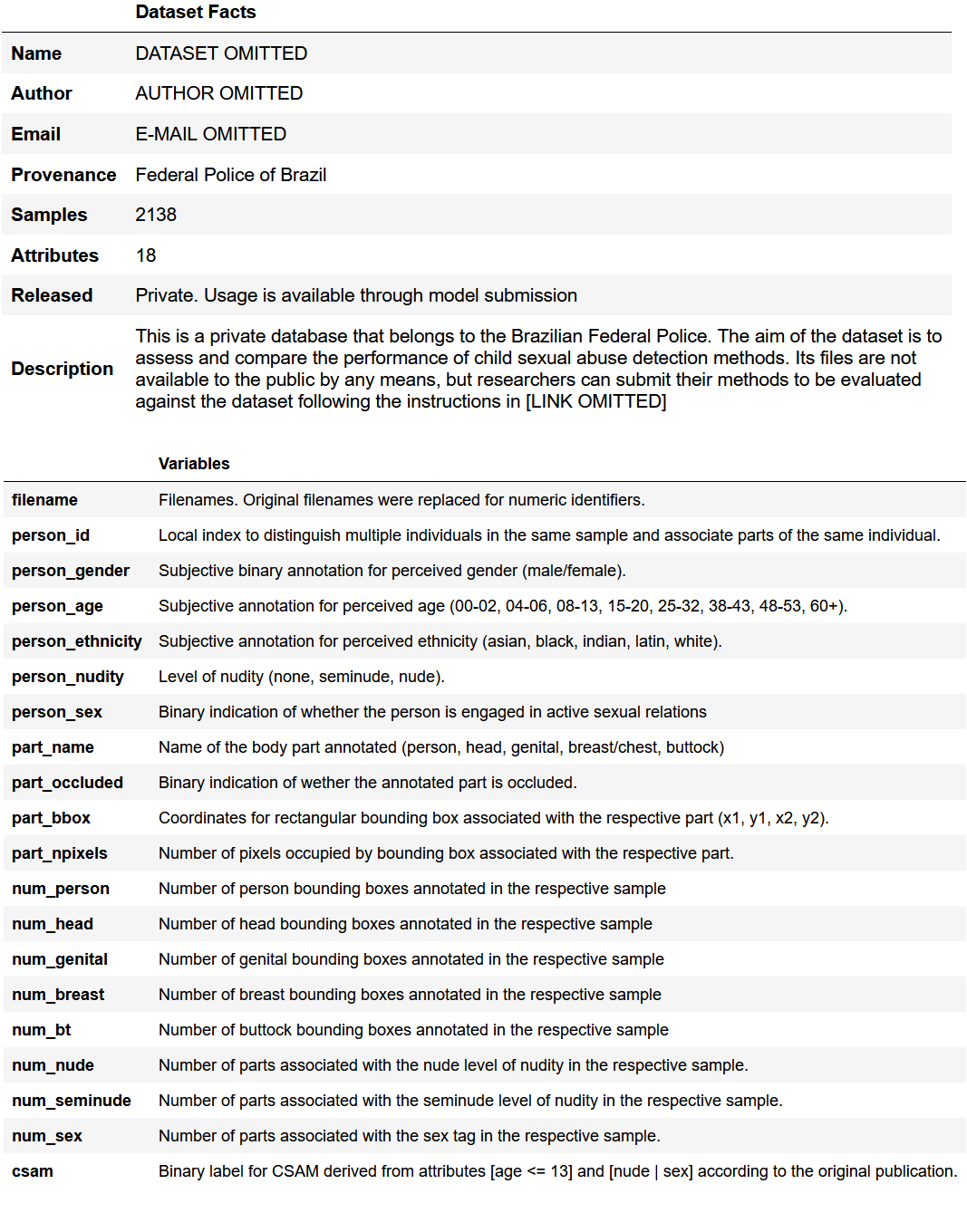}
    \caption{Dataset Facts and Attributes.}
    \label{fig:nlfacts}
\end{figure}

\begin{figure}
    \centering
    \includegraphics[width=0.9\textwidth]{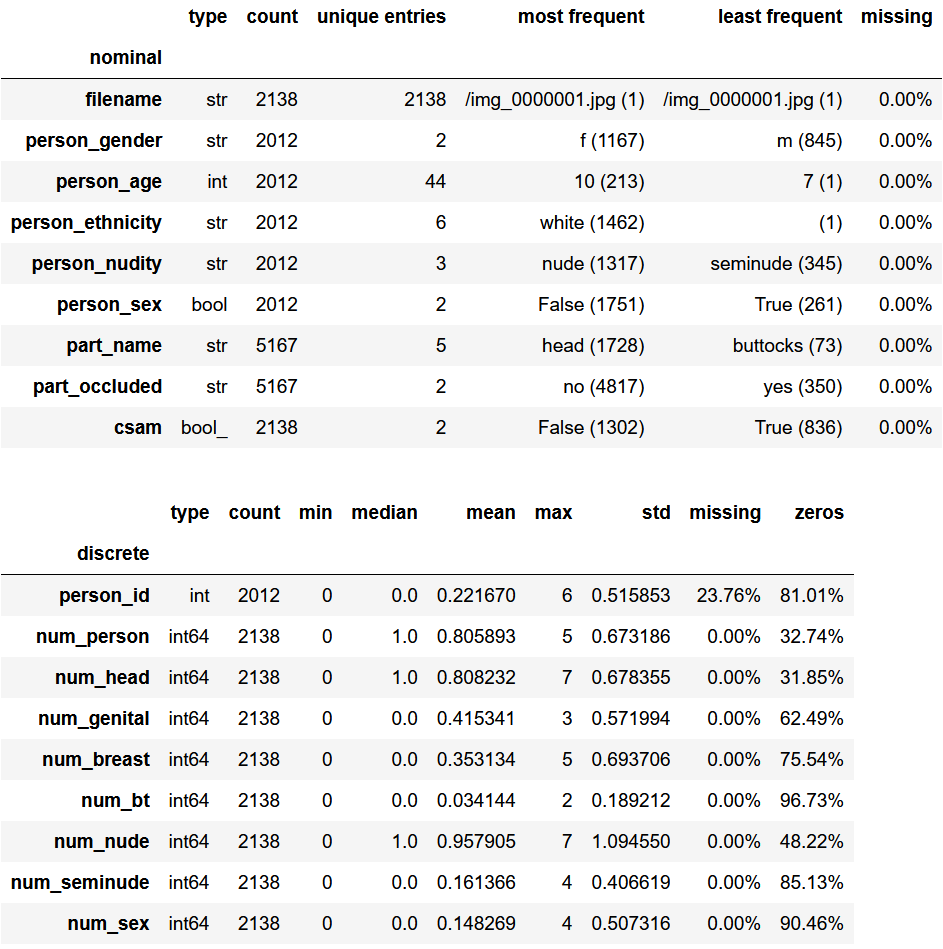}
    \caption{Summary statistics. Note that missing values for attribute person\_id indicate the absence of people in the respective samples. }
    \label{fig:nlsummary}
\end{figure}

\newpage
\section{RCPD 
-- Visual Summary}\label{app:vissummary}

This section contains a more extensive but not exhaustive set of visualizations for manual labels and automatic attributes assigned to RCPD
. Our goal is to provide a comprehensive representation of CSAM datasets both in terms of general statistics and relations between attributes. It requires an interactive interface such that users could submit queries of the desired attributes to relate; thus, visualizations below may contain interactive resources. 

\subsection{Labels}
\begin{figure}[H]
    \centering
    \includegraphics[width=\textwidth]{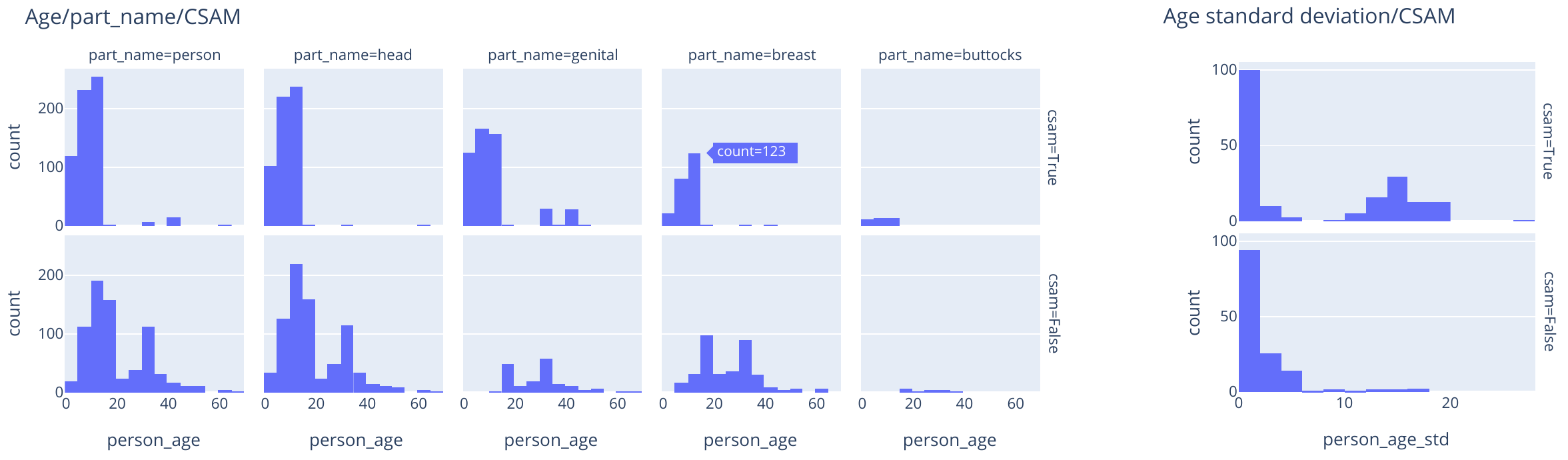}
    \caption{Demonstrating disaggregated plots by showing distributions of age and age standard deviation disaggregated by other dimensions. Interactivity allows inspecting fine-grained information on demand, as demonstrated by the text balloon.}
    \label{fig:app:disag_age}
\end{figure}

In Fig.~\ref{fig:app:rcpd-labels} we provide distributions for all labels from RCPD
, previously listed in Appendix \ref{app:nutrition}. We also demonstrate how disaggregated visualizations can provide important insights into the data. Fig.~\ref{fig:app:disag_age} shows two examples, the first with age distributions disaggregated by two dimensions, labeled parts, and CSAM categories, while the distribution of age standard deviation is split only by CSAM. From the latter figure, we notice that from the few images containing more than one person, negative CSAM samples usually depict people of similar ages. 

\begin{figure}[H]
    \centering
    \includegraphics[width=0.9\textwidth]{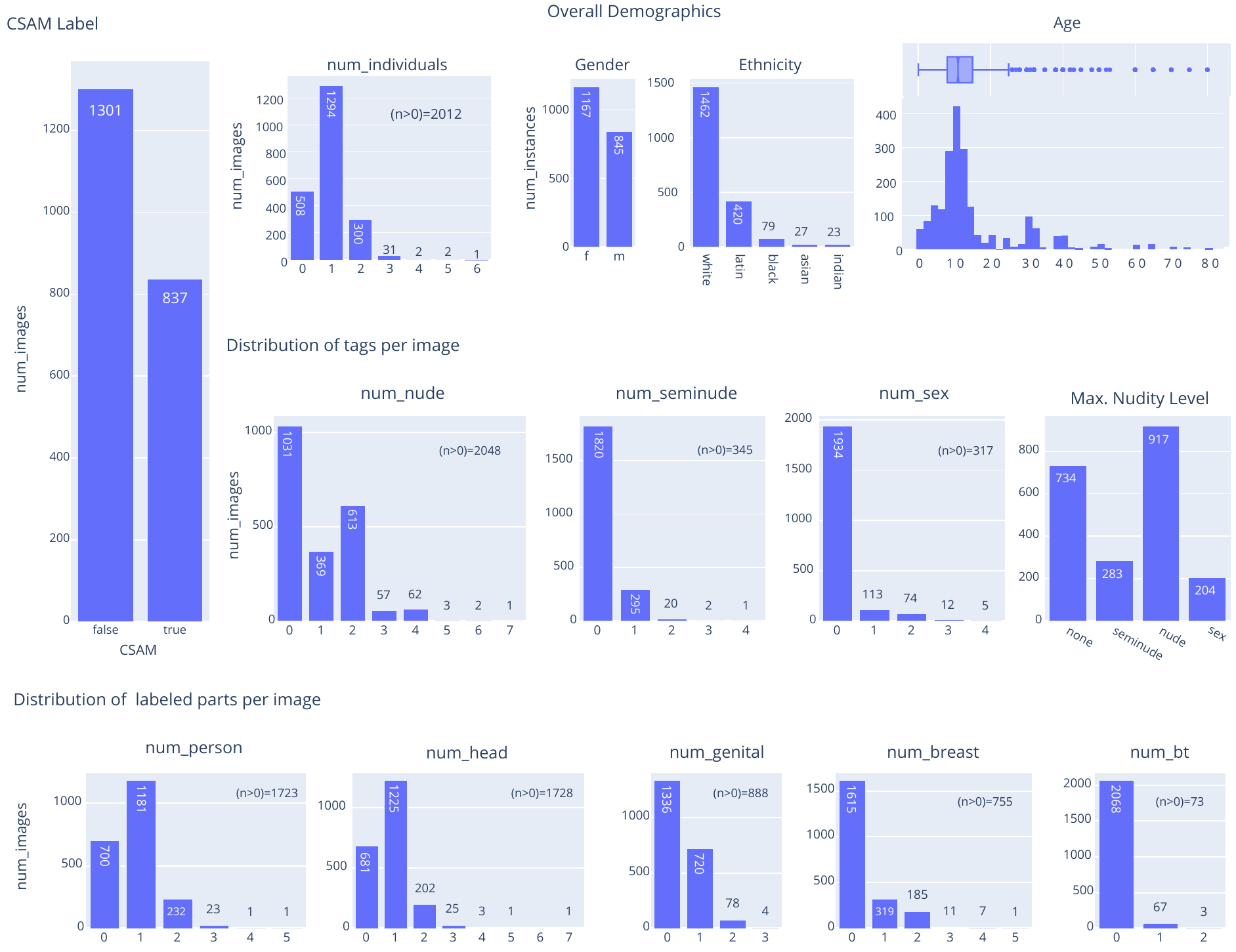}
    \caption{Overview of labels from RCPD
    . Visualizations representing number of instances also provide a total count of instances for number of occurrences $n>0$.}
    \label{fig:app:rcpd-labels}
\end{figure}

\subsection{Extracted attributes}
We present distributions for all features listed in Table \ref{tab:att}. Since we carefully curate the set of attributes, we can craft specialized visualizations for each attribute.  

\begin{figure}[H]
    \centering
    \includegraphics[width=0.8\textwidth]{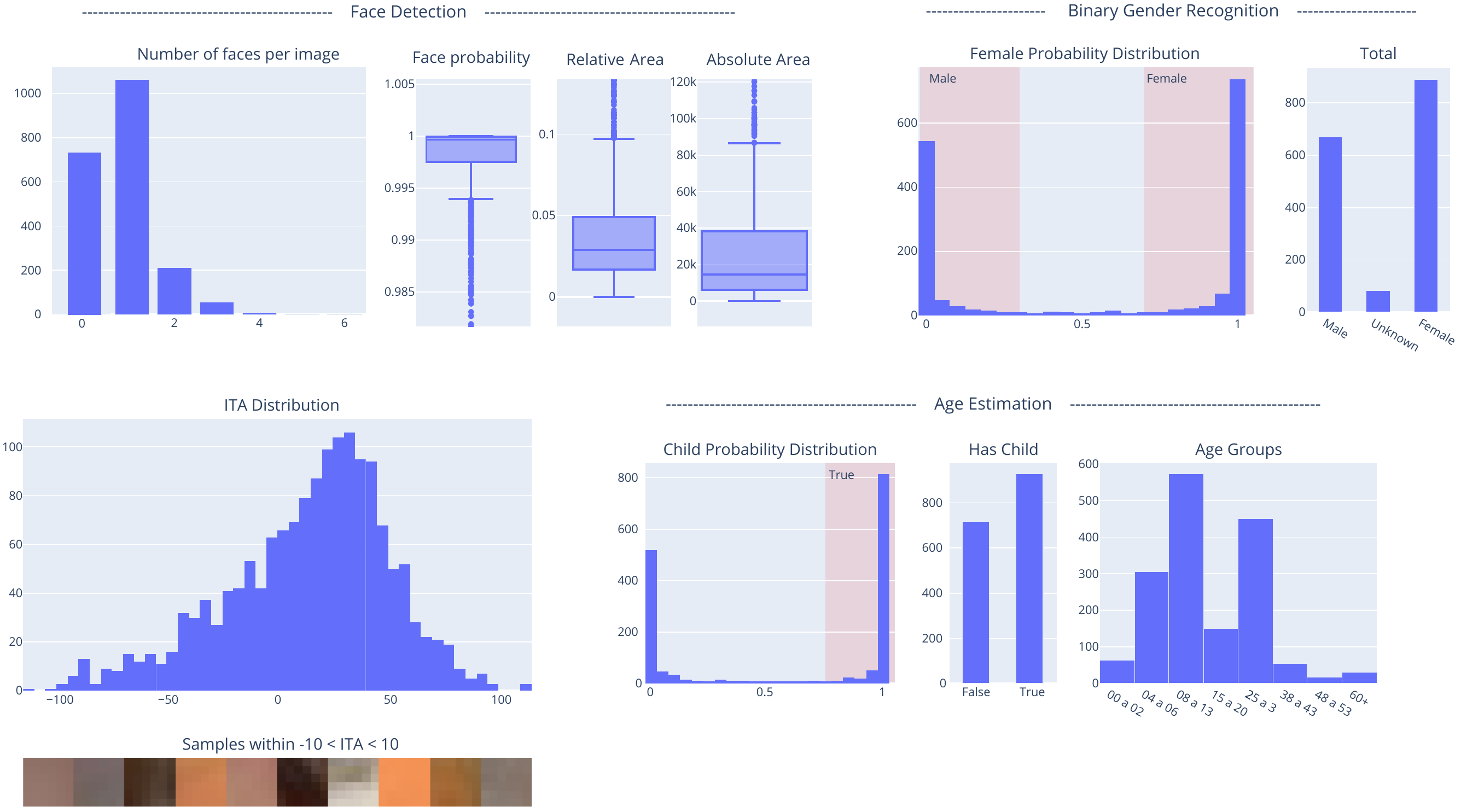}
    \caption{Overview of demographic attributes extracted from RCPD
    . Whenever probabilities are visualized, the threshold applied for classification is represented as a red rectangle bounded by the threshold interval. ITA distribution is accompanied by patch samples from the database representing any chosen ITA interval.}
    \label{fig:my_label}
\end{figure}

\begin{figure}[H]
    \centering
    \includegraphics[width=\textwidth]{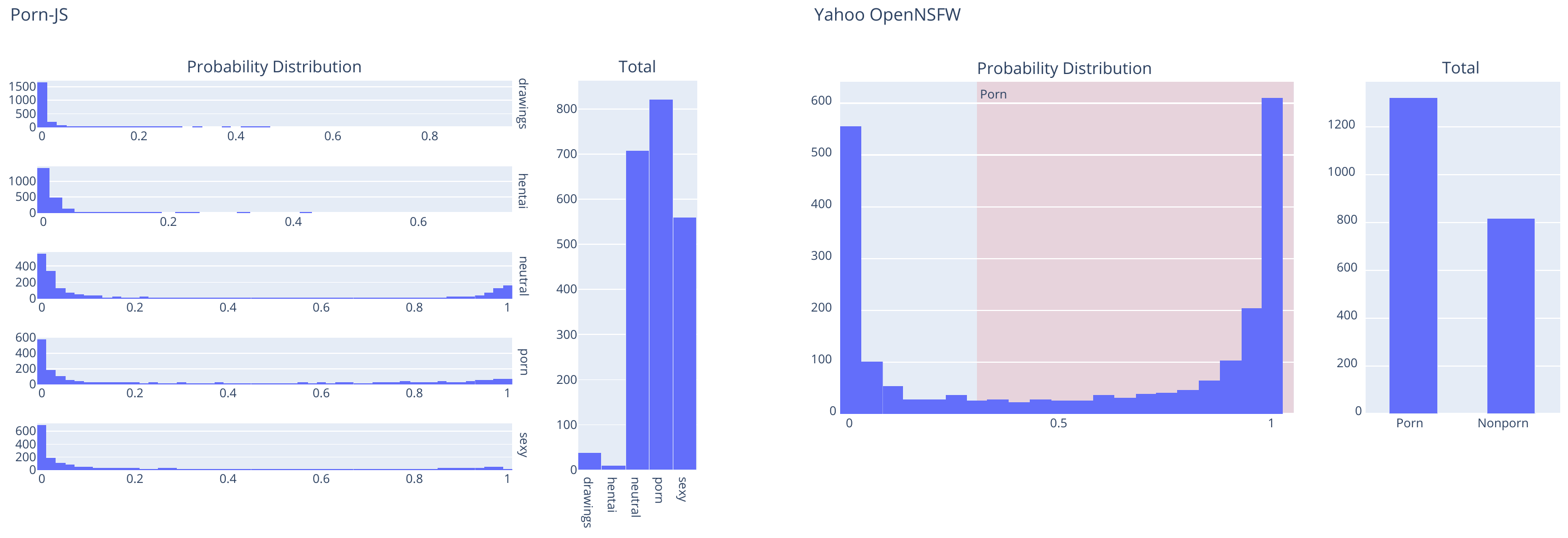}
    \caption{Overview of pornography attributes extracted from RCPD
    . There is no default threshold to visualize for multiclass labels, where classification is performed through maximum activation.}
    \label{fig:my_label}
\end{figure}

\begin{figure}[H]
    \centering
    \includegraphics[width=0.95\textwidth]{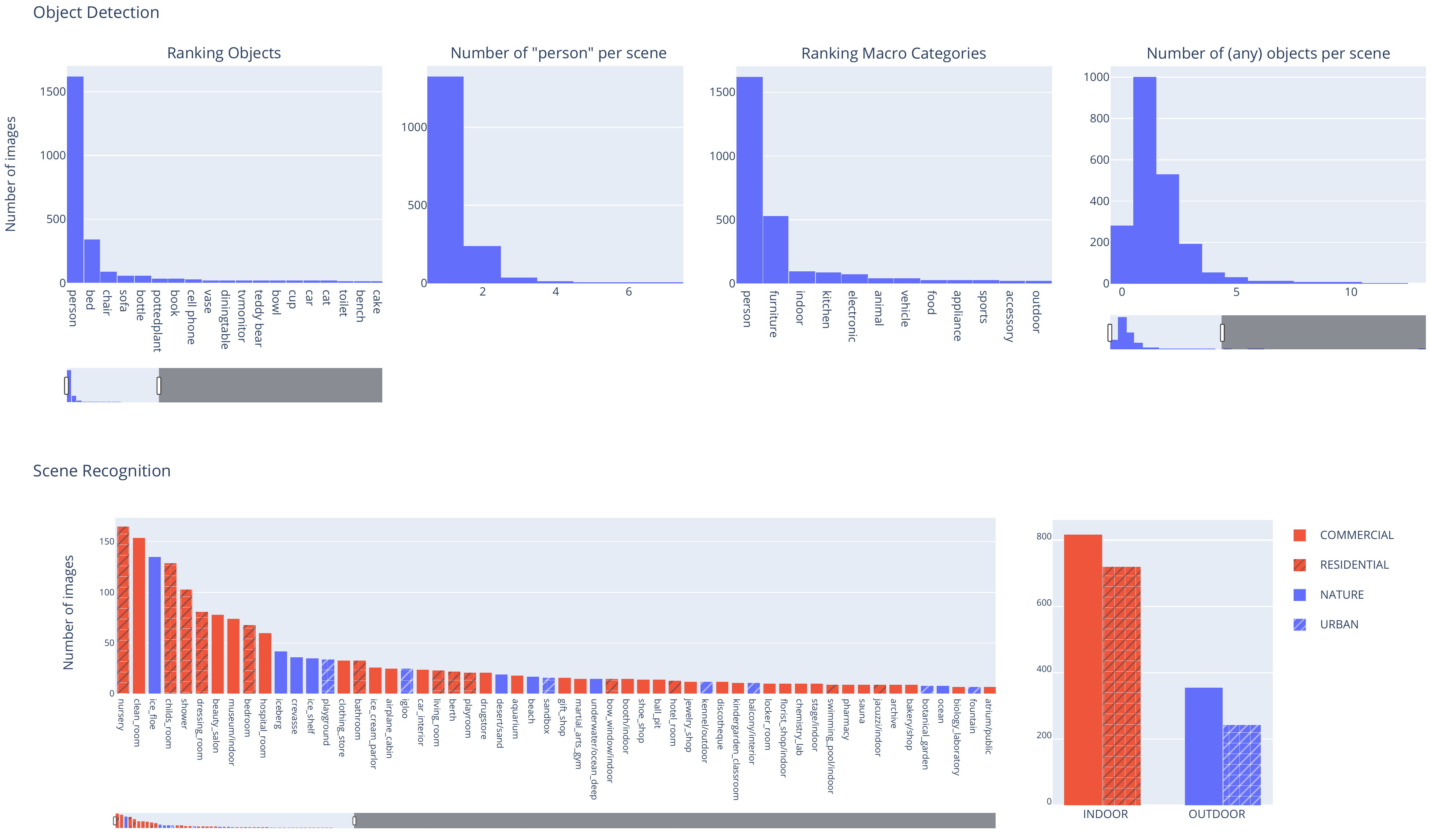}
    \caption{Overview of context attributes extracted from RCPD
    . For histograms with over 20 bins, as is the case for object categories (80 classes) and scenes (365 classes), we add a range slider below the visualization for interactivity.}
    \label{fig:my_label}
\end{figure}

\begin{figure}[H]
    \centering
    \includegraphics[width=0.74\textwidth]{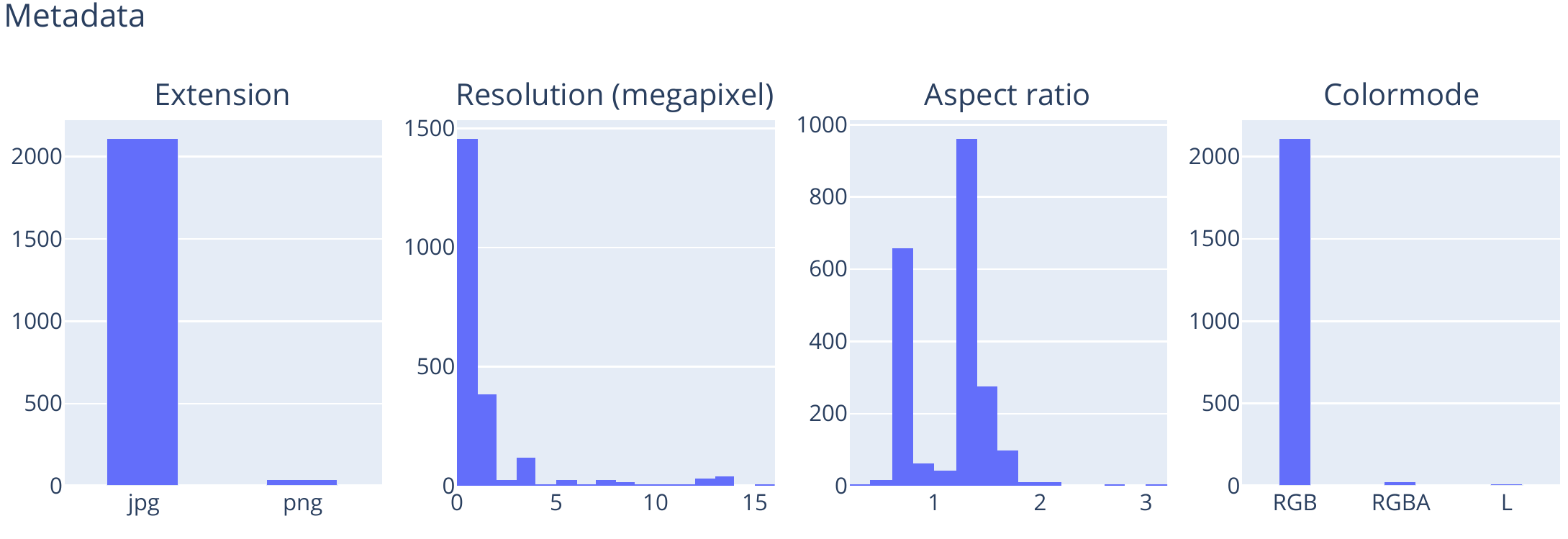}
    \caption{Overview of metadata extracted from RCPD
    .}
    \label{fig:my_label}
\end{figure}

\begin{figure}[H]
    \centering
    \includegraphics[width=0.58\textwidth]{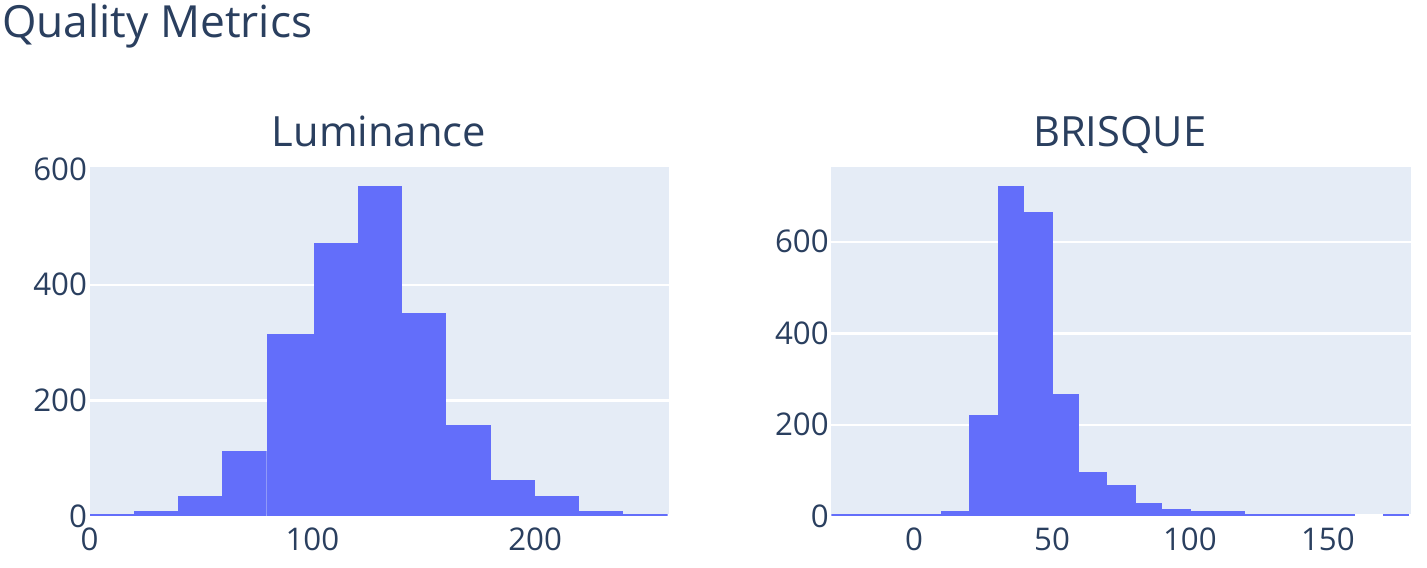}
    \caption{Overview of quality attributes extracted from RCPD
    .}
    \label{fig:my_label}
\end{figure}









\end{document}